\documentclass{article}

\usepackage{microtype}
\usepackage{graphicx}
\usepackage{tabularx}
\usepackage{subcaption}  
\usepackage{booktabs} 
\usepackage{rotating}
\usepackage{adjustbox}
\usepackage{tikz}
\usepackage{float}
\usepackage{natbib}

\usepackage{enumitem}

\usepackage{hyperref}

\usepackage[accepted]{icml2023}

\usepackage{amsmath}
\usepackage{amssymb}
\usepackage{mathtools}
\usepackage{amsthm}

\usepackage[framemethod=TikZ]{mdframed}

\usepackage[capitalize,noabbrev]{cleveref}

\theoremstyle{plain}
\newtheorem{theorem}{Theorem}[section]
\newtheorem{proposition}[theorem]{Proposition}
\newtheorem{lemma}[theorem]{Lemma}
\newtheorem{corollary}[theorem]{Corollary}
\theoremstyle{definition}
\newtheorem{definition}[theorem]{Definition}

\theoremstyle{remark}

\usepackage{thmtools}
\declaretheoremstyle[%
  spaceabove=-6pt,%
  spacebelow=6pt,%
  headfont=\normalfont\itshape,%
  postheadspace=1em,%
  qed=\qedsymbol%
]{mystyle} 
\declaretheorem[name={Proof},style=mystyle,unnumbered,
]{prf}

\newenvironment{claim}[1]{\par\noindent\underline{Claim:}\space#1}{}
\newenvironment{claimproof}[1]{\par\noindent\underline{Proof:}\space#1}{\hfill $\#$}

\usepackage[disable,textsize=tiny]{todonotes}

\icmltitlerunning{Geometric Autoencoders -- What You See is What You Decode}

\DeclareMathOperator{\dkl}{KL}

\DeclareMathOperator*{\argmin}{arg\,min}
\DeclareMathOperator{\tr}{tr}
\DeclareMathOperator{\diag}{diag}
\DeclareMathOperator{\Var}{Var}
\DeclareMathOperator*{\Varl}{Var}
\DeclareMathOperator*{\El}{\mathbb{E}}
\DeclareMathOperator{\ELU}{ELU}

\begin{document}

\twocolumn[
\icmltitle{Geometric Autoencoders -- What You See is What You Decode}



\icmlsetsymbol{equal}{*}

\begin{icmlauthorlist}
\icmlauthor{Philipp Nazari}{hd}
\icmlauthor{Sebastian Damrich}{hd,tb}
\icmlauthor{Fred A. Hamprecht}{hd}
\end{icmlauthorlist}

\icmlaffiliation{hd}{HCI/IWR at University of Heidelberg, 69120 Heidelberg, Germany}
\icmlaffiliation{tb}{University of T\"{u}bingen, 72074 T\"{u}bingen, Germany}

\icmlcorrespondingauthor{Philipp Nazari}{philipp.nazari@gmail.com}
\icmlcorrespondingauthor{Sebastian Damrich}{sebastian.damrich@uni-tuebingen.de}
\icmlcorrespondingauthor{Fred Hamprecht}{fred.hamprecht@iwr.uni-heidelberg.de}

\icmlkeywords{Machine Learning, ICML, Autoencoder, Regularization, Geometry, Dimensionality Reduction, Visualization}

\vskip 0.3in
]



\printAffiliationsAndNotice{}  

\begin{abstract}
Visualization is a crucial step in exploratory data analysis. One possible approach is to train an autoencoder with low-dimensional latent space. Large network depth and width can help unfolding the data. However, such expressive networks can achieve low reconstruction error even when the latent representation is distorted. To avoid such misleading visualizations, we propose first a differential geometric perspective on the decoder, leading to insightful diagnostics for an embedding's distortion, and second a new regularizer mitigating such distortion. Our ``Geometric Autoencoder'' avoids stretching the embedding spuriously, so that the visualization captures the data structure more faithfully. It also flags areas where little distortion could not be achieved, thus guarding against misinterpretation. 

\end{abstract}
\begin{figure*}[ht]
\vskip 0.2in
\begin{center}
     \centering

     \begin{subfigure}[b]{\textwidth}
        \begin{mdframed}[roundcorner=10pt, userdefinedwidth=\textwidth]
        \centering

        \caption{Autoencoders are useful}    
        \label{fig:one-earth}

         \begin{subfigure}[b]{0.49\textwidth}
             \begin{subfigure}[b]{\textwidth}
                \centering
                     \begin{tikzpicture}[font=\scriptsize]
                        \node[minimum width=\textwidth] {Embeddings};
                     \end{tikzpicture}
             \end{subfigure}
         
             \begin{subfigure}[b]{0.23\textwidth}
                \centering
                     \begin{tikzpicture}[font=\tiny]
                        \node[minimum width=\textwidth] {Vanilla AE};
                     \end{tikzpicture}
             \end{subfigure}
             \begin{subfigure}[b]{0.23\textwidth}
                \centering
                 \begin{tikzpicture}[font=\tiny]
                    \node[minimum width=\textwidth] {\textbf{Geometric AE}};
                 \end{tikzpicture}
             \end{subfigure}
             \begin{subfigure}[b]{0.23\textwidth}
                \centering
                     \begin{tikzpicture}[font=\tiny]
                        \node[minimum width=\textwidth] {$t$-SNE};
                     \end{tikzpicture}
             \end{subfigure}
             \begin{subfigure}[b]{0.23\textwidth}
                \centering
                     \begin{tikzpicture}[font=\tiny]
                        \node[minimum width=\textwidth] {PCA};
                     \end{tikzpicture}
             \end{subfigure}

             \begin{subfigure}[b]{0.23\textwidth}
                 \centering
                 \includegraphics[width=\linewidth]{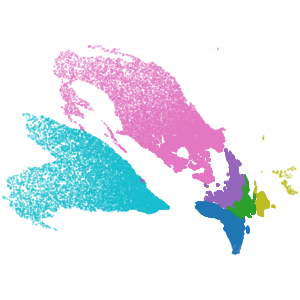}
             \end{subfigure}
             \begin{subfigure}[b]{0.23\textwidth}
                 \centering
                 \includegraphics[width=\linewidth]{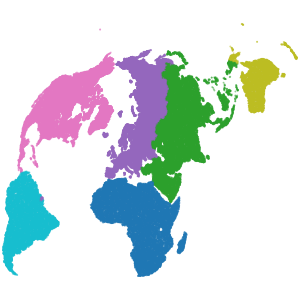}
             \end{subfigure}
             \begin{subfigure}[b]{0.23\textwidth}
                 \centering
                 \includegraphics[width=\linewidth]{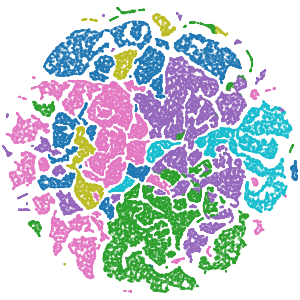}
             \end{subfigure}
             \begin{subfigure}[b]{0.23\textwidth}
                 \centering
                 \includegraphics[width=\linewidth]{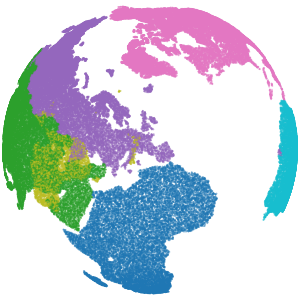}
             \end{subfigure}

            \begin{subfigure}[b]{\textwidth}
               \centering
                    \begin{tikzpicture}[font=\tiny]
                       \node[minimum width=\textwidth] {};
                    \end{tikzpicture}
            \end{subfigure}

         \end{subfigure}
         \vrule
         \hfill
         \begin{subfigure}[b]{0.49\textwidth}

             \begin{subfigure}[b]{\textwidth}
                \centering
                     \begin{tikzpicture}[font=\scriptsize]
                        \node[minimum width=\textwidth] {Diagnostics};
                     \end{tikzpicture}
             \end{subfigure}
            
             \begin{subfigure}[b]{0.23\textwidth}
                \centering
                     \begin{tikzpicture}[font=\tiny]
                        \node[minimum width=\textwidth] {Vanilla AE};
                     \end{tikzpicture}
             \end{subfigure}
             \begin{subfigure}[b]{0.23\textwidth}
                \centering
                 \begin{tikzpicture}[font=\tiny]
                    \node[minimum width=\textwidth] {\textbf{Geometric AE}};
                 \end{tikzpicture}
             \end{subfigure}
             \begin{subfigure}[b]{0.23\textwidth}
                \centering
                     \begin{tikzpicture}[font=\tiny]
                        \node[minimum width=\textwidth] {Vanilla AE};
                     \end{tikzpicture}
             \end{subfigure}
             \begin{subfigure}[b]{0.23\textwidth}
                \centering
                 \begin{tikzpicture}[font=\tiny]
                    \node[minimum width=\textwidth] {\textbf{Geometric AE}};
                 \end{tikzpicture}
             \end{subfigure}

             \begin{subfigure}[b]{0.23\textwidth}
                 \centering
                 \includegraphics[width=\linewidth]{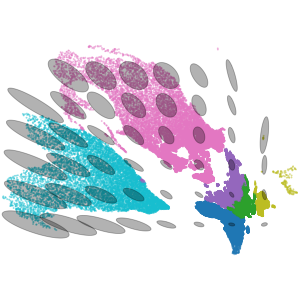}
             \end{subfigure}
             \begin{subfigure}[b]{0.23\textwidth}
                 \centering
                 \includegraphics[width=\linewidth]{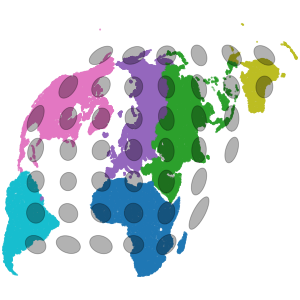}
             \end{subfigure}
             \begin{subfigure}[b]{0.23\textwidth}
                 \centering
                 \includegraphics[width=\linewidth]{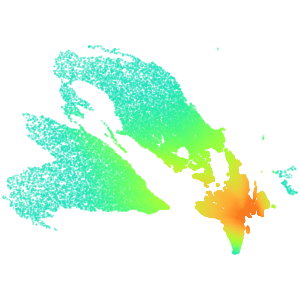}
             \end{subfigure}
             \begin{subfigure}[b]{0.23\textwidth}
                 \centering
                 \includegraphics[width=\linewidth]{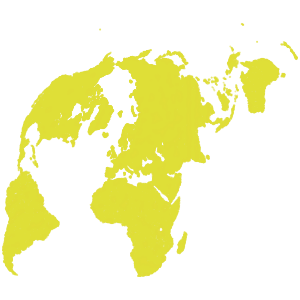}
             \end{subfigure}
             
             \begin{subfigure}[b]{0.47\textwidth}
                \centering
                     \begin{tikzpicture}[font=\tiny]
                        \node[minimum width=\textwidth] {};
                     \end{tikzpicture}
             \end{subfigure}
            \begin{subfigure}[b]{0.47\textwidth}
                \centering
                 \begin{subfigure}[b]{\textwidth}
                     \centering
                     \includegraphics[width=\linewidth]{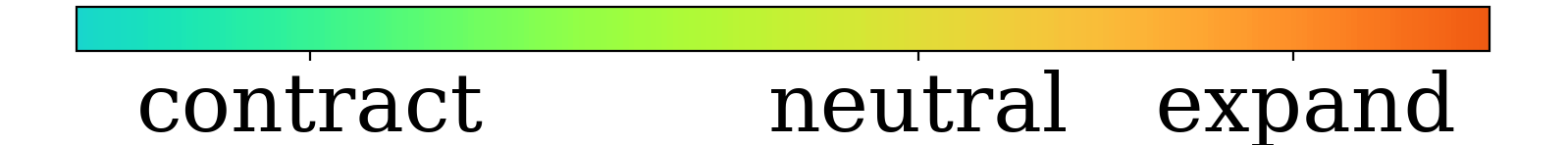}
                 \end{subfigure}
            \end{subfigure}

         \end{subfigure}

        \end{mdframed}
     \end{subfigure}

     \begin{subfigure}[b]{0.60\textwidth}
        \begin{mdframed}[roundcorner=10pt, userdefinedwidth=\textwidth]
        \centering
        \caption{Geometric vs. Vanilla Autoencoder}
        \label{fig:geom_vs_vanilla}

        
         \begin{subfigure}[b]{0.05\textwidth}
            \centering
                 \begin{tikzpicture}[font=\tiny]
                    \node[rotate=90] {};
                 \end{tikzpicture}
         \end{subfigure}
         \begin{subfigure}[b]{0.18\textwidth}
            \centering
                 \begin{tikzpicture}[font=\tiny]
                    \node[minimum width=\textwidth] {MNIST};
                 \end{tikzpicture}
        \end{subfigure}
         \begin{subfigure}[b]{0.18\textwidth}
            \centering
                 \begin{tikzpicture}[font=\tiny]
                    \node[minimum width=\textwidth] {FashionMNIST};
                 \end{tikzpicture}
         \end{subfigure}
         \begin{subfigure}[b]{0.18\textwidth}
            \centering
                 \begin{tikzpicture}[font=\tiny]
                    \node[minimum width=\textwidth] {Zilionis};
                 \end{tikzpicture}
         \end{subfigure}
         \begin{subfigure}[b]{0.18\textwidth}
            \centering
                 \begin{tikzpicture}[font=\tiny]
                    \node[minimum width=\textwidth] {PBMC};
                 \end{tikzpicture}
         \end{subfigure}
         \begin{subfigure}[b]{0.18\textwidth}
            \centering
             \begin{tikzpicture}[font=\tiny]
                \node[minimum width=\textwidth] {CElegans};
             \end{tikzpicture}
         \end{subfigure}

         \begin{subfigure}[b]{0.05\textwidth}
            \centering
                 \begin{tikzpicture}[font=\tiny]
                    \node(1)[minimum width=3.6\textwidth, rotate=90] {\textbf{Geometric AE}};
                 \end{tikzpicture}
         \end{subfigure}
         \begin{subfigure}[b]{0.18\textwidth}
             \centering
             \includegraphics[width=\linewidth]{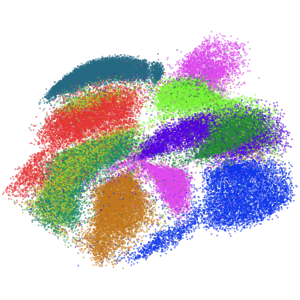}
         \end{subfigure}
         \begin{subfigure}[b]{0.18\textwidth}
             \centering
             \includegraphics[width=\linewidth]{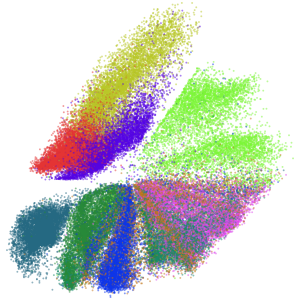}
         \end{subfigure}
         \begin{subfigure}[b]{0.18\textwidth}
             \centering
             \includegraphics[width=\linewidth]{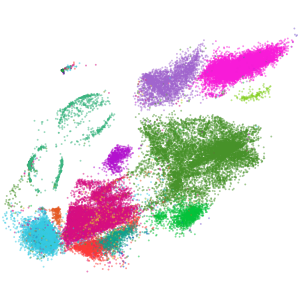}
         \end{subfigure}
         \begin{subfigure}[b]{0.18\textwidth}
             \centering
             \includegraphics[width=\linewidth]{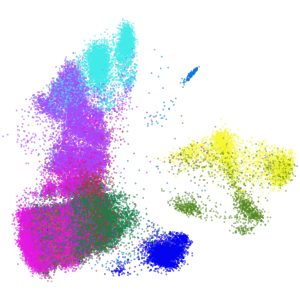}
         \end{subfigure}
         \begin{subfigure}[b]{0.18\textwidth}
             \centering
             \includegraphics[width=\linewidth]{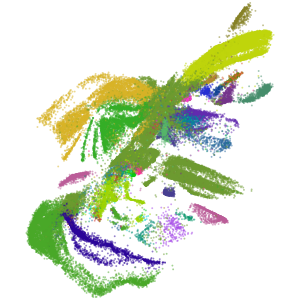}
         \end{subfigure}

         \begin{subfigure}[b]{0.05\textwidth}
            \centering
                 \begin{tikzpicture}[font=\tiny]
                    \node(1)[minimum width=3.6\textwidth, rotate=90] {Vanilla AE};
                 \end{tikzpicture}
         \end{subfigure}
         \begin{subfigure}[b]{0.18\textwidth}
             \centering
             \includegraphics[width=\linewidth]{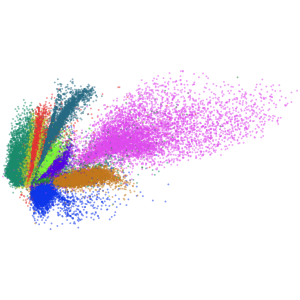}
         \end{subfigure}
         \begin{subfigure}[b]{0.18\textwidth}
             \centering
             \includegraphics[width=\linewidth]{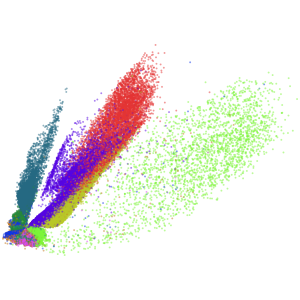}
         \end{subfigure}
         \begin{subfigure}[b]{0.18\textwidth}
             \centering
             \includegraphics[width=\linewidth]{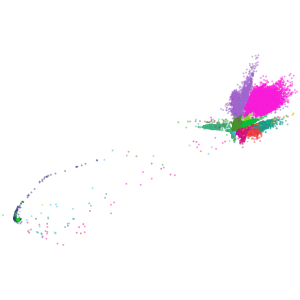}
         \end{subfigure}
         \begin{subfigure}[b]{0.18\textwidth}
             \centering
             \includegraphics[width=\linewidth]{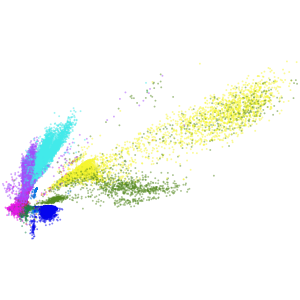}
         \end{subfigure}
         \begin{subfigure}[b]{0.18\textwidth}
             \centering
             \includegraphics[width=\linewidth]{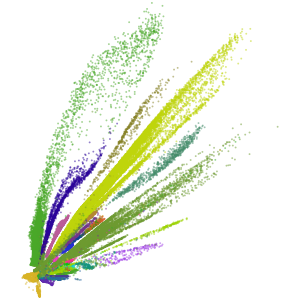}
         \end{subfigure}
         \end{mdframed}
     \end{subfigure}
     \hfill
     \begin{subfigure}[b]{0.39\textwidth}
     \begin{mdframed}[roundcorner=10pt, userdefinedwidth=\textwidth]
         \centering
         \caption{Geometric vs. UMAP Autoencoder}
         \label{fig:geom_vs_umap_ae}
         \vspace{1.25mm}

        \begin{subfigure}{0.9\textwidth}
             \begin{subfigure}[b]{0.07\textwidth}
                \centering
                 \begin{tikzpicture}[font=\tiny]
                    \node[rotate=90] {};
                 \end{tikzpicture}
             \end{subfigure}
             \begin{subfigure}[b]{0.29\textwidth}
                \centering
                     \begin{tikzpicture}[font=\tiny]
                        \node[minimum width=\textwidth] {Embedding};
                     \end{tikzpicture}
             \end{subfigure}
             \begin{subfigure}[b]{0.29\textwidth}
                \centering
                     \begin{tikzpicture}[font=\tiny]
                        \node[minimum width=\textwidth] {Indicatrices};
                     \end{tikzpicture}
             \end{subfigure}
             \begin{subfigure}[b]{0.29\textwidth}
                \centering
                     \begin{tikzpicture}[font=\tiny]
                        \node[minimum width=\textwidth] {Determinant};
                     \end{tikzpicture}
             \end{subfigure}
    
             \begin{subfigure}[b]{0.07\textwidth}
                \centering
                     \begin{tikzpicture}[font=\tiny]
                        \node(1)[minimum width=4.14\textwidth, rotate=90, align=center] {\textbf{Geometric AE}};
                     \end{tikzpicture}
             \end{subfigure}
             \begin{subfigure}[b]{0.29\textwidth}
                 \centering
                 \includegraphics[width=\linewidth]{media/pbmc/geomreg/latents.png}
             \end{subfigure}
             \begin{subfigure}[b]{0.29\textwidth}
                 \centering
                 \includegraphics[width=\linewidth]{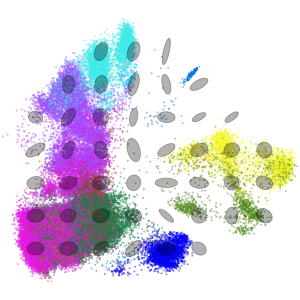}
             \end{subfigure}
             \begin{subfigure}[b]{0.29\textwidth}
                 \centering
                 \includegraphics[width=\linewidth]{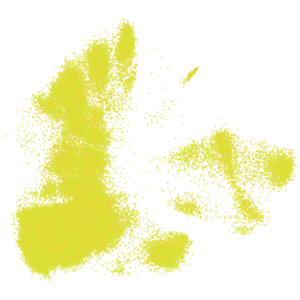}
             \end{subfigure}
             \begin{subfigure}[b]{0.07\textwidth}
                \centering
                     \begin{tikzpicture}[font=\tiny]
                        \node(1)[minimum width=4.14\textwidth, rotate=90, align=center] {UMAP AE};
                     \end{tikzpicture}
             \end{subfigure}
             \begin{subfigure}[b]{0.29\textwidth}
                 \centering
                 \includegraphics[width=\linewidth]{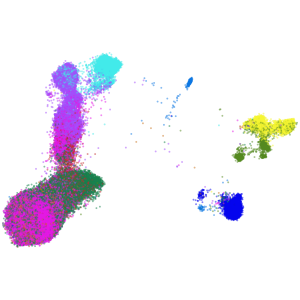}
             \end{subfigure}
             \begin{subfigure}[b]{0.29\textwidth}
                 \centering
                 \includegraphics[width=\linewidth]{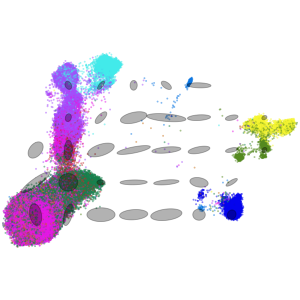}
             \end{subfigure}
             \begin{subfigure}[b]{0.29\textwidth}
                 \centering
                 \includegraphics[width=\linewidth]{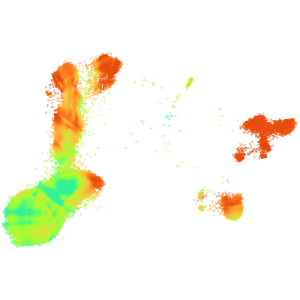}
             \end{subfigure}
        \end{subfigure}
        \begin{subfigure}{0.07\textwidth}
            \centering
            \includegraphics[width=\linewidth]{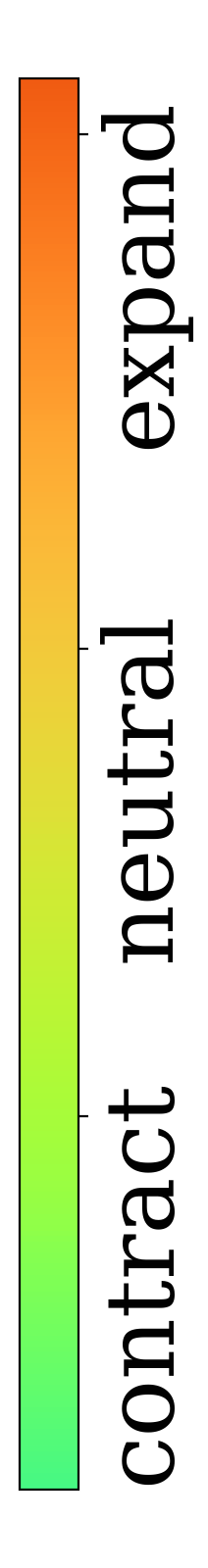}
        \end{subfigure}
        \vspace{2.1mm}
        \end{mdframed}
     \end{subfigure}
    \caption{Panel~(\subref{fig:one-earth}) provides an example for the usefulness of autoencoders for visualization. While autoencoders are able to unravel the Earth dataset, $t$-SNE disregards global structure, and PCA projects Eurasia onto Australia. The geometric autoencoder ensures that the relative sizes of the continents are much better preserved than by the vanilla autoencoder. We use diagnostics, indicatrices and a determinant plot, to demonstrate that our geometric autoencoder's embedding is more faithful, contracting more homogeneously. Panel~(\subref{fig:geom_vs_vanilla}) shows how our geometric regularizer improves upon a vanilla autoencoder on a number of datasets. Panel~(\subref{fig:geom_vs_umap_ae}) compares our geometric autoencoder with a UMAP autoencoder on the PBMC dataset. The diagnostics show that the UMAP decoder expands much more inhomogeneously than our proposed method, distorting the relative cluster sizes.}
\label{fig:figure-one}
\end{center}
\vskip -0.2in
\end{figure*}

\section{Introduction}
\label{introduction}

The acquisition of larger and more complex datasets -- with a dimensionality of a few thousand, for example in bio-informatics~\cite{pbmc, zilionis, celegans} -- has boosted the development of recent machine learning algorithms.
While such high dimensionality allows for encoding an increasing amount of information, it also makes human interpretation more difficult. A common method for exploring high-dimensional datasets is two- or three-dimensional visualization.

Today's state-of-the-art algorithms for dimensionality reduction are UMAP~\cite{umap, umap-true} and $t$-SNE~\cite{tsne, tsne-acc, tsne-single-cell}. Both UMAP and $t$-SNE tend to preserve local structure, which gives them great unfolding power, at the expense of rendering global structure faithfully (see Figure~\ref{fig:one-earth}).

In contrast to standard UMAP and $t$-SNE, autoencoders~\cite{hinton2006reducing} find representations that afford approximate reconstruction of the original, high-dimensional dataset. They also allow the embedding of additional measurements after training.
If the autoencoder is linear, it reduces to PCA~\cite{pca, pca-ae, pca-ae-2}, see also Appendix~\ref{sec:appendix-pca}. Non-linear autoencoders have much greater representational power.
In Figure~\ref{fig:one-earth} we show that autoencoders can produce meaningful maps of the globe where PCA (projecting New Zealand onto Italy) and $t$-SNE (distorting land mass beyond recognition) both fail.

While non-linearity enables autoencoders to unfold the data,
it can also hinder the interpretability of an autoencoder's latent representation. Powerful encoders can introduce distortions that equally powerful decoders can resolve, leaving the reconstruction loss unaffected.
The result is a misleading embedding with near perfect reconstruction, subverting the idea that a 2D latent space forces only the most salient features to be visualized.
In deep networks this defect can be amplified to the point that even a simple dataset, such as the \textit{Earth} dataset (see Appendix~\ref{sec:appendix-datasets}) in Figure~\ref{fig:one-earth}, can become hardly recognizable. In the vanilla autoencoder's embedding, South and North America each seem to be bigger than Eurasia and Africa combined. Nevertheless, the autoencoder achieves low reconstruction loss, because the decoder contracts the Americas while expanding the rest of the world, so that the reconstruction accurately reflects the continents' actual sizes.

We take an intuitive geometric approach to different methods of measuring how encoder and decoder introduce distortion in the embedding, which assist the practitioner in understanding a given embedding. 
To make sure that what we see in the embedding is closer to what the decoder reconstructs, and thus to the structure of the dataset, we propose taming the decoder's geometric properties by constructing a regularizer that pushes the decoder towards being area-preserving. This approach is similar to the one in~\citet{lee2022regularized} which encourages the decoder to be a scaled isometry -- a more restrictive property than area-preservation.

The decoder maps the low-dimensional latent space to the high-dimensional output space. If, for example, we consider a two-dimensional latent space and a three-dimensional ambient space, the autoencoder learns to place a curved surface into $\mathbb R^3$ which, in a suitable measure, best approximates the higher dimensional dataset.
Intuitively, one can visualize the decoder's task as fitting a surface into output-space, stretching it arbitrarily. While some stretching might be necessary to approximate the dataset well, excessive stretching introduces unnecessary distortions in latent space.
Loosely speaking, the geometric regularizer makes the surface resist stretching intrinsically.

We propose visualizing the decoder's expansion, which we can think of as the surface's stretching, by a heatmap of the \textit{generalized Jacobian determinant} (closely related to the ``Riemannian Volumeform''~\cite{LeeSmoothManifolds} and to the work of~\citet{chen2018metrics}) and by \textit{indicatrices}~\cite{tissot, tensor-glyph-warping}. While the generalized Jacobian determinant measures the decoder's undirected contraction, indicatrices additionally show its anisotropy. Their size and elongation enable the practitioner to more faithfully interpret any autoencoder, see Figures~\ref{fig:one-earth},~\subref{fig:geom_vs_umap_ae}. We further endow latent-space with a \textit{pullback metric} which allows us to measure and mitigate the decoder's variance in contraction.\newline

To sum up, we propose diagnostic tools for visualizing local distortion of two-dimensional autoencoders and construct a geometric regularizer reducing those distortions, leading to a more faithful embedding.
We provide the code as an open-source package for PyTorch. It can be found at \url{https://github.com/hci-unihd/GeometricAutoencoder}.

\section{Preliminaries}
\label{preliminaries}
\subsection{Problem Setting}
\label{sec:problem-setting}
Throughout this work, we assume that there is a dataset $X$ living in some high-dimensional Euclidean space $\mathbb R^n$.
We view an autoencoder as a concatenation of two functions $\mathbb R^n \xrightarrow[]{E} \mathbb R^l \xrightarrow[]{D} \mathbb R^n$,
where $E$ is the encoder, $D$ the decoder and $l < n$ is the dimensionality of latent space. Both decoder and encoder are realized as (deep) neural networks, which are jointly trained to minimize the $\ell_2$ loss between the dataset and its reconstruction. Under some mild assumptions (Section~\ref{limitations}), the decoder's image defines an $l$-dimensional manifold $M$ (the ``reconstruction manifold'') living in $\mathbb R^n$ with an atlas consisting of only a single chart, the decoder's inverse $D^{-1} \colon M \to \mathbb R^l$. The encoder $E \colon \mathbb R^n \to \mathbb R^l$ can be seen as placing an input point onto the codomain of the chart. The decoder thus defines the manifold on which the autoencoder can place reconstructions. The encoder specifies the position on the manifold by outputting the position on the global chart. During training, updating the decoder changes the reconstruction manifold, while updating the encoder changes the position on the chart and thus on the manifold.

If an autoencoder was trained to optimal reconstruction loss, it would essentially project data points orthogonally to the reconstruction manifold, see Appendix~\ref{sec:appendix-orthogonal-projection}. Even in this case, the encoder could still locally stretch and contract the embedding as long as the decoder undoes these distortions.
A priori, this is not visible in the embedding space. For example, the vanilla autoencoder's embedding of the Earth dataset in Figure~\ref{fig:one-earth} disproportionally expands the Americas. Despite this visual distortion, the reconstruction loss is half that of the geometric autoencoder. This is only possible if the decoder contracts the enlarged embedding of the Americas again.
We will present a way of measuring such avoidable contraction and ultimately mitigating it as much as possible.

It is known from multivariate calculus that the Jacobian determinant of a continuously differentiable function $f \colon \mathbb R^n \to \mathbb R^n$ at a point $p \in \mathbb R^n$ measures how $f$ transforms an infinitesimal volume centered at $p$.
In order to develop the concept of a Jacobian determinant for the decoder, which in general acts between spaces of different dimensionality, we first generalize the ordinary case to smooth immersions $F \colon M \to N$, smooth maps with injective differential everywhere, between manifolds. This requires some machinery which we introduce in the following paragraph.

\subsection{A Note on Differential Geometry}
\label{sec:differential-geometry}
In this section we introduce basic concepts from differential geometry. For a detailed treatment, see ~\citet{LeeRiemannianManifolds, LeeSmoothManifolds}.

One of the core concepts from differential geometry is that of a \textit{(smooth) manifold}, a space that locally looks like Euclidean Space; it can be covered by open sets $U$, each of which is homeomorphic to an open subset of $\mathbb R^n$. Such a homeomorphism is called a \textit{chart}. Furthermore, we require the transition maps between charts to be diffeomorphisms.\newline
The directions of a manifold $M$ at a point $p \in M$ are captured by the \textit{tangent space} $T_pM$ at $p$, which can be thought of as the best linear approximation of the manifold.\newline
A smooth map $f \colon M \to N$ between two manifolds can be linearly approximated around each point $p \in M$. This approximation is called the \textit{differential} of $f$ at $p$, denoted by $d_pf$, and maps from the tangent space corresponding to $p$ to that of its image, $d_pf \colon T_pM \to T_{f(p)}N$. In coordinates, it is given by the Jacobian matrix $J_pF$ of $F$ at $p$.\newline
Distances and angles at a point $p$ of a Riemannian manifold are determined by the \textit{metric tensor}, a bilinear, positive definite map acting on the tangent space at $p$.
The Euclidean metric tensor $g_e$ is given by the Euclidean inner product.

\subsection{The Generalized Jacobian Determinant}
\label{sec:generalized-jacobian-determinant}
In this section we generalize the concept of a Jacobian determinant to smooth immersions $F \colon M \to N$ between manifolds of dimension $m$ and $n$, from which the ordinary case emerges as a special case.

Assume $(N, g)$ to be a Riemannian manifold, then $F$ induces a volume form on $F(M)$ in the following way:
\begin{proposition}
\label{prop:volume-form}
    Assume $F$ is a diffeomorphism onto its image and $M$ is oriented.
    Then there exists a volume form $\omega_g$ on $F(M)$, the Riemannian volume form, 
    which in the smooth oriented coordinates $x_1,...,x_l$ induced by $F$ is given by \mbox{$\omega_g = \sqrt{\det\left[\left( J_pF \right)^t J_pF\right]} dx_1 \wedge ... \wedge dx_l$}.
\end{proposition}
\begin{prf}
Use Proposition 15.6 and 15.31 in~\citet{LeeSmoothManifolds}.
\end{prf}
The square-root factor of the Riemannian volume form shows how volumes are changed locally by $F$, and can thus be seen as a generalization of the Jacobian determinant. We call its square the \textit{generalized Jacobian determinant}, which captures information about the distortion of angles and directed stretching. This gives rise to the ``pullback metric``, which we introduce in the next paragraph.

\subsection{The Pullback Metric}
\label{pullback-metric}
In order to faithfully interpret the latent space of an autoencoder, it is crucial to know how angles and distances would appear after decoding. 
This can be achieved by equipping latent space with a metric tensor that measures angles and distances as they would be mapped to the output manifold. The resulting metric tensor on latent space is the \textit{pullback metric}~\cite{LeeSmoothManifolds}. See Figure~\ref{fig:pullback} for an illustration.

To construct the pullback metric tensor, we endow ambient space with the Euclidean metric $g_e$, making it a Riemannian Manifold $(\mathbb R^n, g_e)$. An immersion $F \colon \mathbb R^l \to \mathbb R^n$, which we will later choose to be the decoder $D$, induces a pullback metric $F^*g_e$ on its domain in the following way:
Given a point $p \in \mathbb R^l$ and two tangent vectors $v, w \in T_p \mathbb R^l$, their inner product in the pullback metric is defined as the inner product of their images under the decoder's differential,
\begin{equation}\label{eq:pullback-metric-abstract}
F^*g_{e_p}(v, w) \coloneqq g_{e_F(p)}(d_pFv, d_pFw),
\end{equation}
where $d_pF \colon T_p \mathbb R^l \to T_{F(p)} \mathbb R^n$ is the differential of $F$ at $p$.

In coordinates, this pullback metric takes a very simple form, just depending on the Jacobian of $F$:
\begin{proposition}[Pullback Metric in Coordinates]
\label{prop:pullback-metric-coordinates}
The pullback of $g_e$ under $F$ at $p \in \mathbb R^l$ is in coordinates given by
\begin{equation}
    \label{eq:pullback-metric-coordinates}
    \langle \cdot, \cdot \rangle_p \coloneqq F^*g_{e_p} = \left( J_pF \right)^t J_pF \in \mathbb R^{l, l},
\end{equation}
where $J_pF \in \mathbb R^{n \times l}$ is the Jacobian matrix of $F$ at $p$.
\end{proposition}
\begin{prf}
See Appendix~\ref{sec:appendix-coordinates}.
\end{prf}

Equation~\eqref{eq:pullback-metric-coordinates} indicates the connection between the pullback metric and the generalized Jacobian determinant introduced in Section~\ref{sec:generalized-jacobian-determinant}. Indeed, the pullback metric measures lengths in latent space as lengths along the immersed manifold.

\subsection{Indicatrices}
\label{method-indicatrices}
While the generalized Jacobian determinant provides information about the decoder's undirected contraction, it does not tell us anything about its isotropy or directed contraction. Therefore, we propose visualizing the pullback metric tensor fields using ``indicatrices``~\cite{tissot, tensor-glyph-warping}. Consider a smooth immersion $F$ between two manifolds $M$ and $N$, as in the setting of Section~\ref{sec:generalized-jacobian-determinant}.
\begin{definition}[Indicatrix]
An \textit{indicatrix} at a point $p \in M$ is the unit sphere in the pullback metric induced by $F$ at $p$.
\end{definition}
Since the differential linearly approximates $F$ at $p$, we may think of an indicatrix as consisting of those points around $p$ which $F$ approximately maps to a unit sphere around $F(p)$. An indicatrix centered at $p$ thus tells us which directions are squeezed and which are expanded. It makes distortions originating from the function $F$ visible.
A set of indicatrices, distributed over the dataset, allows to identify regions which are contracted or expanded as well as the direction of the stretching. See Figure~\ref{fig:indicatrices-concept} for a visual explanation.

\subsection{Application to Autoencoders}
\label{application-to-autoencoders}
For the geometric autoencoder, we equip the decoder's image with the restriction of the Euclidean metric $g_e$, which we then pull back using the decoder $D$. In particular, all the above applies to the special case where $F=D$ is the decoder.
For limitations of our method, see Section~\ref{limitations}.

\section{Geometric Autoencoders}
\label{geometric-autoencoders}
Unregularized autoencoders tend to contract the embedding inhomogeneously. In this section, we discuss diagnostics for this distortion, as well as a regularizer mitigating the variance in contraction. See Figure~\ref{fig:figure-one} for an overview.

\subsection{Diagnostics}
\label{diagnostics}

\subsubsection{Generalized Jacobian Determinant}
\label{diagnostics-determinant}
To prevent misinterpreting an embedding due to inhomogeneous contraction of the decoder, we propose highlighting areas in latent space based on the generalized Jacobian determinant, which we plot as a heat map on the embedding as opposed to the background shading in~\citet{chen2018metrics}. This helps interpreting embeddings more faithfully: In Figure~\ref{fig:one-earth}, the determinant plot reveals that the heavily clustered data lies in an area which the decoder expands. Thus, one can infer that Europe, Russia and Africa combined are not actually smaller than each of the two Americas.

\subsubsection{Indicatrices}
\label{diagnostics-indicatrices}

\begin{figure*}[ht]
\vskip 0.2in
\begin{center}
     \begin{subfigure}[b]{0.39\textwidth}        
         \includegraphics[width=\textwidth]{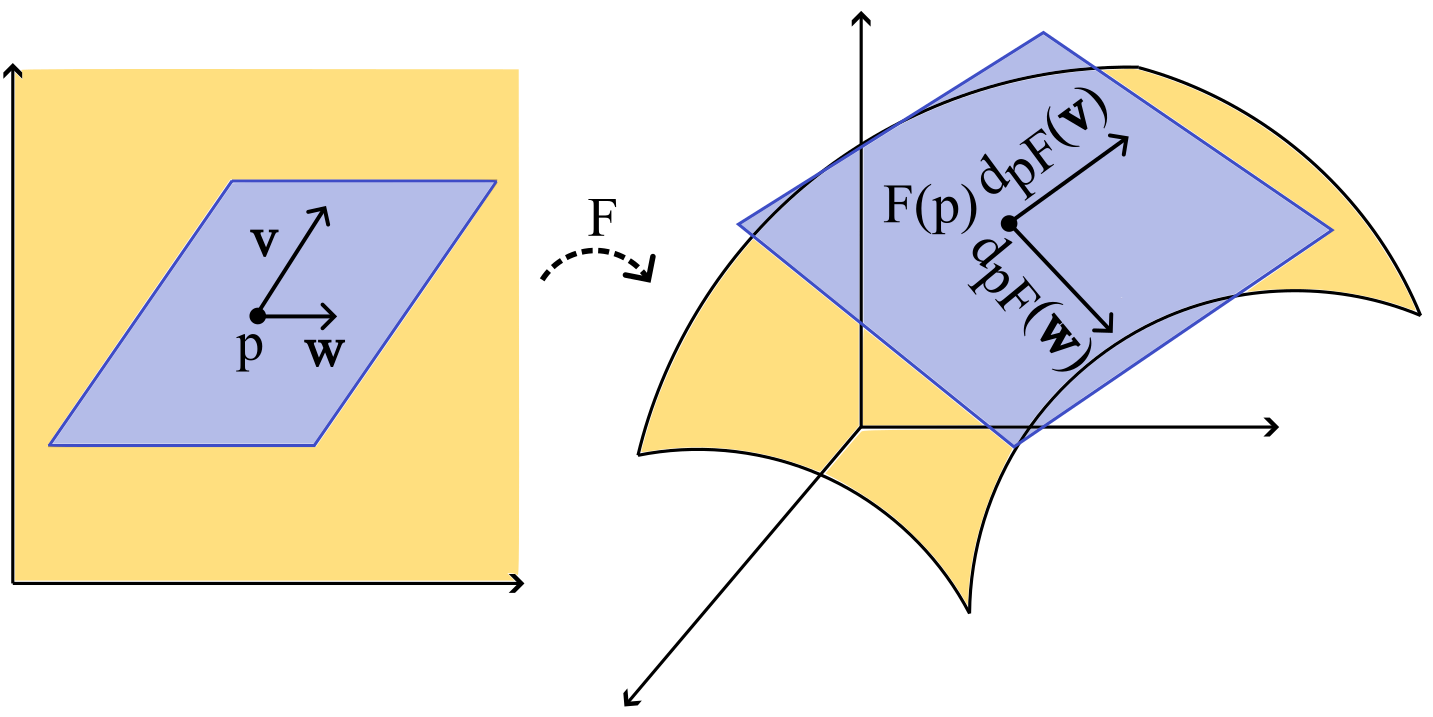}
     \caption{Pullback Metric (Concept)}
     \label{fig:pullback}
     \end{subfigure}
     \hspace{10mm}
     \begin{subfigure}[b]{0.39\textwidth}        
         \includegraphics[width=\textwidth]{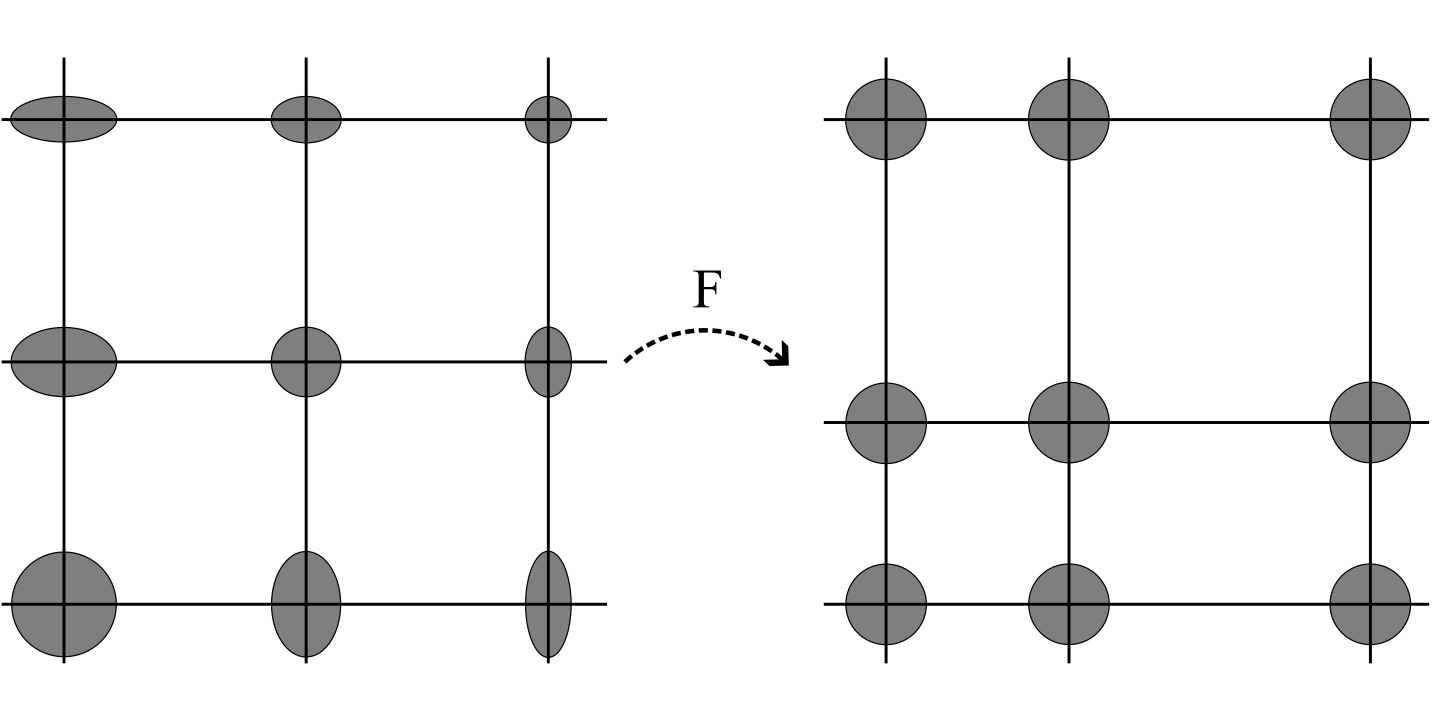}
     \caption{Indicatrices (Concept)}
     \label{fig:indicatrices-concept}
     \end{subfigure}
     \caption{
     Panel~(\subref{fig:pullback}) illustrates the pullback metric along a smooth map $F \colon \mathbb R^2 \to \mathbb R^3$. Given a point $p \in \mathbb R^2$ and two vectors $u$ and $v$ from the tangent space at $p$ (purple), their product in the pullback metric is defined as the product of their images under $F$'s differential $d_pF$ (on the right). While $v$ and $w$ are not orthonormal in the euclidean metric, they are orthonormal in the pullback metric, since their images are.
     Panel~(\subref{fig:indicatrices-concept}) illustrates indicatrices. Consider a map $F \colon \mathbb R^2 \to \mathbb R^2$, $(x, y) \mapsto (x^2, y^2)$ which distorts a regular grid as displayed. The shape of the indicatrices makes this distortion visible in the input space. 
     }
\end{center}
\vskip -0.2in
\end{figure*}

Given a decoder $D\colon \mathbb R^l \to \mathbb R^n$, we approximate the indicatrix at $p \in \mathbb R^l$ by the convex hull of the family of vectors $v_i / \sqrt{\langle v_i, v_i \rangle_p}$, where the $v_i$ are sampled uniformly from the Euclidean unit circle at $p$. 
As a visualization technique, we plot for a given point $p$ in latent space the convex hull of the vectors $v_i$ as a patch around $p$. See Figure~\ref{fig:indicatrices-concept} for an example. The points $p$ are chosen as a regular grid in the convex hull of the embedding. The inhomogeneity of the embedding is reflected in the indicatrices. The variance of the decoder's undirected contraction is indicated by the variance of the indicatrices' volumes. For example, in Figure~\ref{fig:one-earth} the vanilla decoder's indicatrices on Europe are smaller than those on North America (Figure~\ref{fig:one-earth}). The decoder's directed contraction is encoded by the shape of a single indicatrix; for the vanilla decoder's in Figure~\ref{fig:one-earth} most of them are elongated towards Europe, indicating that the decoder has to contract in that direction in order to reconstruct the dataset. A locally isotropic decoder, for example the one of PCA, has round indicatrices (Figure~\ref{fig:indicatrices}, column 4).

\subsection{Regularization}
\label{regularization}
We discussed above how the generalized Jacobian determinant measures the local contraction and expansion of a decoder. A faithful embedding avoids any stretching unnecessary for reconstruction. Therefore, it is natural to regularize the decoder to have uniform generalized Jacobian determinant.
To achieve that, we calculate the generalized Jacobian determinant at every embedding point in the embedding of a minibatch $B$ and calculate the variance of their logarithm. This defines our regularizer $\mathcal L_{\det}$,
\begin{equation}
    \label{eq:geomreg-objective}
    \mathcal L_{\det} = \Var_{x \sim \mathcal U(B)} \left[ \log \left( \det \left( J_{E(x)}D \right)^t J_{E(x)}D \right) \right].
\end{equation}
The total loss amounts to $\mathcal L \to \mathcal L_{\text{rec}} + \alpha \mathcal L_{\det}$, where $\alpha$ is a hyperparameter controlling the importance of the regularizer compared to the usual reconstruction loss $\mathcal L_{\text{rec}}$, typically the mean squared error.
In Appendix~\ref{sec:appendix-gradients} we explain why gradients resulting from the regularizer are propagated through both the encoder and the decoder. 

Computing the variance of logarithms ensures that the autoencoder cannot minimize the secondary objective by globally expanding the embedding:
\begin{lemma}[Scale Invariance of the Regularizer]
\label{lem:scale_inv}
If the decoder scales with a factor $\beta \in \mathbb R \setminus \{0\}$, the objective $\mathcal L_{\det}$ stays invariant.
\end{lemma}
\begin{prf}
See Appendix~\ref{sec:appendix-invariance}.
\end{prf}

\section{Related Work}
\label{related-work}

Since the invention of autoencoders~\cite{ae} and their application to visualization, see e.g. ~\citet{hinton2006reducing}, numerous regularizations have been proposed to avoid over-fitting.

Two popular strategies are contractive autoencoders~\cite{cae} and denoising autoencoders~\cite{dae}. Both are geared towards classification, rather than visualization, as they aim to produce locally constant embeddings. Sparse autoencoders~\cite{ng2011sparse, makhzani2014k} are regularized to have sparse hidden activations instead of compressing to a bottleneck dimension, making them unsuitable for visualization as well. Therefore, we omitted these three classical regularized autoencoders from our quantitative evaluation. 

Variational autoencoders~\cite{variational-bias, vae} are tailored towards generating samples from a prespecified prior distribution in latent space, typically a Gaussian. As a result, the embeddings are usually densely packed together to allow smooth interpolation. This is not ideal for visualization, especially of clustered datasets.
\citet{rae} suggest to replace the variational framework with various deterministic regularizers. However, none of them directly address the geometric properties of the embedding. Furthermore,~\citet{rae} do not consider visualization.

More similar to our method are topological autoencoders~\cite{topoae}. Effectively, they encourage the encoder to preserve the minimum spanning tree of the dataset. Recently, \citet{trofimov2023learning} proposed to regularize autoencoders based on a more refined method for comparing the topology between point clouds. Instead of regularizing the topology of the embedding, our proposed method addresses the geometry. Other works in this area have tried to turn the decoder into an isometry, a map that preserves pairwise distances on a local scale. The Markov-Lipschitz autoencoder~\cite{li2020markov} directly regularizes local distances and across several layers of the network. Isometric autoencoders~\cite{gropp2020isometric} try to achieve isometry of the decoder instead by preserving the norm of Monte-Carlo sampled unit vectors in latent space under multiplication with the decoder's Jacobian. The work of~\citet{chen2020learning} regularizes the pullback metric tensor directly via a Frobenius norm, but their coordinate-dependent measure has a bias for decoders with Jacobian of small norm~\cite{lee2022regularized}. Instead, ~\citet{lee2022regularized} propose a different regularization of the pullback metric that induces the decoder to become a scaled isometry. Our approach aligns closely with that of~\citet{lee2022regularized}, but is less restrictive and only encourages the decoder to become area-preserving, see Appendix~\ref{sec:appendix-relation-lee}. 

Other regularized autoencoders include neighborhood reconstructing autoencoders~\cite{lee2021neighborhood} that try to reconstruct neighborhoods of data points by local approximation of the decoder. The geometry regularized autoencoders of~\citet{duque2022geometry} regularize the latent layer to stay close to a previously computed embedding, e.g., a neighbor embedding with UMAP. Our method does not employ neighborhood relations, but only uses the reconstruction of individual point and our geometric regularizer. The recently proposed geometrically regularized autoencoder of~\citet{janggeometrically} extends denoising and contractive autoencoders to the setting where the data and possibly the latent space are known, non-Euclidean Riemannian manifolds. Crucial for visualization, we only consider 2D Euclidean latent spaces. While not explored here, our work readily applies to a general Riemannian data space by pulling back its metric tensor to latent space.

Improving the structure of latent space activations in the supervised setting is an active area of research, too~\citep{zhao2018softmax, scott2021mises}.

The most popular non-parametric dimensionality reduction algorithms are the neighbor embedding methods \mbox{$t$-SNE}~\cite{tsne, tsne-single-cell, tsne-acc} and UMAP~\cite{umap}. Their relation is discussed in~\citet{tsne-umap}. Both $t$-SNE and UMAP usually do not include a decoder. There is, however, a parametric implementation of UMAP, which can be implemented as an autoencoder with UMAP loss on the embedding~\cite{parametric-umap}.
In this setup, our diagnostics revealed that UMAP embeddings have significant variance in local contraction and expansion, see Figures~\ref{fig:determinants} and~\ref{fig:indicatrices}.

We use indicatrices for visualizing the pullback metric. These are related to Tissot indicatrices~\cite{tissot}, commonly used to visualize distortions in world maps, and Tensor Glyphs~\cite{tensor-glyph-warping}. Both of those methods are more complicated than ours which is more in line with the equidistance-lines and -plots in~\citet{chen2018metrics, lee2022regularized}. We recommend plotting indicatrices across the entire embedding instead of only at isolated points. Magnification factor plots have been put forward in~\citet{chen2018metrics} though we suggest restricting them to embedding points. This is meaningful, since judging the area-distortion is most relevant in regions that contain data.

\section{Experiments}
\label{experiments}
\subsection{Experimental Setup}
\label{experimental-setup}
\textbf{Datasets}
Besides the classical image datasets MNIST~\cite{mnist} and FashionMNIST~\cite{fashion-mnist}, we use the three single-cell datasets Zilionis~\cite{zilionis}, PBMC~\cite{pbmc} and CElegans~\cite{celegans}. For illustration only, we generate an \textit{Earth} dataset consisting of points randomly sampled from the unit sphere $S^2 \subset \mathbb R^3$, wherever there would be landmass on earth.
More information can be found in Appendix~\ref{sec:appendix-datasets}.

\textbf{Baselines}
We use UMAP~\cite{umap}, \mbox{$t$-SNE}~\cite{tsne} and PCA~\cite{pca} as baselines, as well as a vanilla autoencoder, an autoencoder with UMAP side-loss~\cite{parametric-umap} and the topological autoencoder~\cite{topoae}. For the former non-parametric models as well as for the UMAP autoencoder's side loss, we use the default parameters. For the topological autoencoder, we weigh the topological loss for all datasets by $\lambda = 0.5$, recommended by~\citet{topoae} for the MNIST dataset.

\textbf{Architecture and Training}
Encoder and decoder of all the autoencoder models have four layers of width $100$, with $\ELU$~\cite{elu} activations.
This architecture is very similar to the standard architecture of~\citet{parametric-umap}, and differs from it only by an additional layer as well as in the activation function.
See Appendix~\ref{sec:appendix-training} for more information about our training procedure. For the proposed geometric autoencoder, we found $\alpha = 0.1$ to be a good weight for the geometric loss term.

\subsection{Evaluation}
\begin{figure*}[ht]
\vskip 0.2in
\begin{center}
     \begin{subfigure}[b]{0.13\textwidth}
        \centering
             \begin{tikzpicture}[font=\tiny]
                \node[minimum width=\textwidth] {Geom AE};
             \end{tikzpicture}
        \end{subfigure}
     \begin{subfigure}[b]{0.13\textwidth}
        \centering
             \begin{tikzpicture}[font=\tiny]
                \node[minimum width=\textwidth] {Vanilla AE};
             \end{tikzpicture}
     \end{subfigure}
     \begin{subfigure}[b]{0.13\textwidth}
        \centering
             \begin{tikzpicture}[font=\tiny]
                \node[minimum width=\textwidth] {Topo AE};
             \end{tikzpicture}
     \end{subfigure}
     \begin{subfigure}[b]{0.13\textwidth}
        \centering
         \begin{tikzpicture}[font=\tiny]
            \node[minimum width=\textwidth] {UMAP AE};
         \end{tikzpicture}
     \end{subfigure}
     \begin{subfigure}[b]{0.13\textwidth}
        \centering
             \begin{tikzpicture}[font=\tiny]
                \node[minimum width=\textwidth] {PCA};
             \end{tikzpicture}
     \end{subfigure}
     \begin{subfigure}[b]{0.13\textwidth}
        \centering
         \begin{tikzpicture}[font=\tiny]
            \node[minimum width=\textwidth] {$t$-SNE};
         \end{tikzpicture}
     \end{subfigure}
     \begin{subfigure}[b]{0.13\textwidth}
        \centering
         \begin{tikzpicture}[font=\tiny]
            \node[minimum width=\textwidth] {UMAP};
         \end{tikzpicture}
     \end{subfigure}

     \begin{subfigure}[b]{0.13\textwidth}
         \centering
         \includegraphics[width=\linewidth]{media/mnist/geomreg/latents.png}
         \caption{}
         \label{fig:embeddings-mnist-geomreg}
     \end{subfigure}
     \begin{subfigure}[b]{0.13\textwidth}
         \centering
         \includegraphics[width=\linewidth]{media/mnist/vanilla/latents.png}
         \caption{}
         \label{fig:main-evaluation-b}
         \label{fig:embeddings-mnist-vanilla}
     \end{subfigure}
     \begin{subfigure}[b]{0.13\textwidth}
         \centering
         \includegraphics[width=\linewidth]{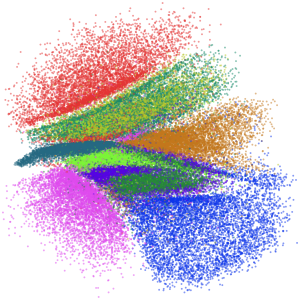}
         \caption{}
         \label{fig:main-evaluation-c}
     \end{subfigure}
     \begin{subfigure}[b]{0.13\textwidth}
         \centering
         \includegraphics[width=\linewidth]{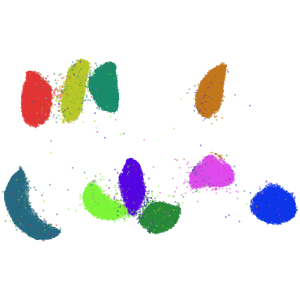}
         \caption{}
     \end{subfigure}
     \begin{subfigure}[b]{0.13\textwidth}
         \centering
         \includegraphics[width=\linewidth]{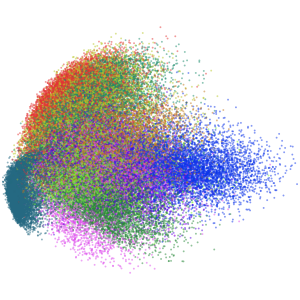}
         \caption{}
     \end{subfigure}
     \begin{subfigure}[b]{0.13\textwidth}
         \centering
         \includegraphics[width=\linewidth]{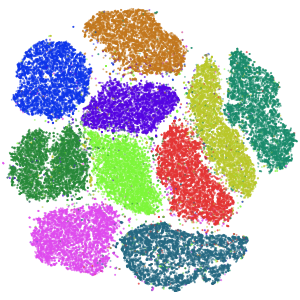}
         \caption{}
     \end{subfigure}
     \begin{subfigure}[b]{0.13\textwidth}
         \centering
         \includegraphics[width=\linewidth]{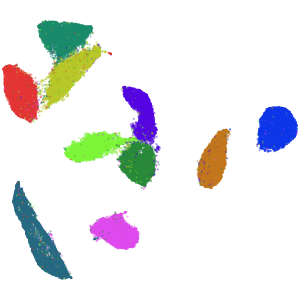}
         \caption{}
     \end{subfigure}

     \begin{subfigure}[b]{0.13\textwidth}
         \centering
         \includegraphics[width=\linewidth]{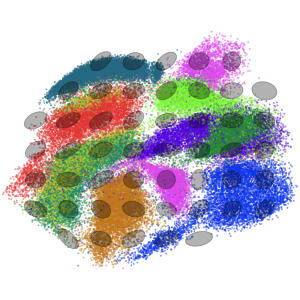}
         \caption{}
		\label{fig:indicatrices-mnist-geomae}
     \end{subfigure}
     \begin{subfigure}[b]{0.13\textwidth}
         \centering
         \includegraphics[width=\linewidth]{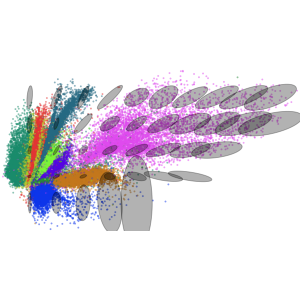}
         \caption{}
		   \label{fig:indicatrices-mnist-vanilla}
     \end{subfigure}
     \begin{subfigure}[b]{0.13\textwidth}
         \centering
         \includegraphics[width=\linewidth]{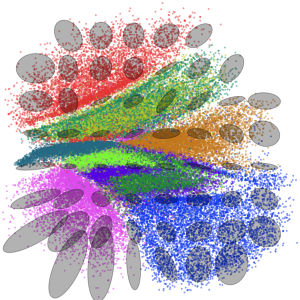}
         \caption{}
		   \label{fig:indicatrices-mnist-topoae}
     \end{subfigure}
     \begin{subfigure}[b]{0.13\textwidth}
         \centering
         \includegraphics[width=\linewidth]{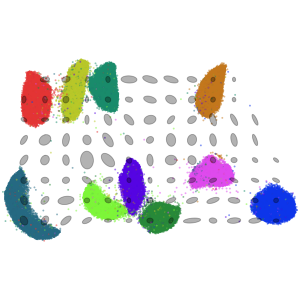}
         \caption{}
		  \label{fig:indicatrices-mnist-pumap}
     \end{subfigure}
     \begin{subfigure}[b]{0.13\textwidth}
         \centering
         \includegraphics[width=\linewidth]{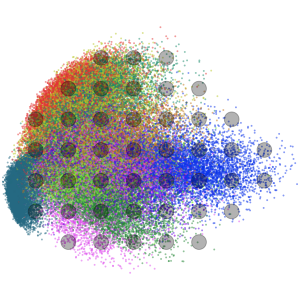}
         \caption{}
         \label{fig:indicatrices-mnist-pca}
     \end{subfigure}
     \begin{subfigure}{0.26\textwidth}
         \begin{subfigure}[b]{0.29\textwidth}
             \centering
             \includegraphics[width=\linewidth]{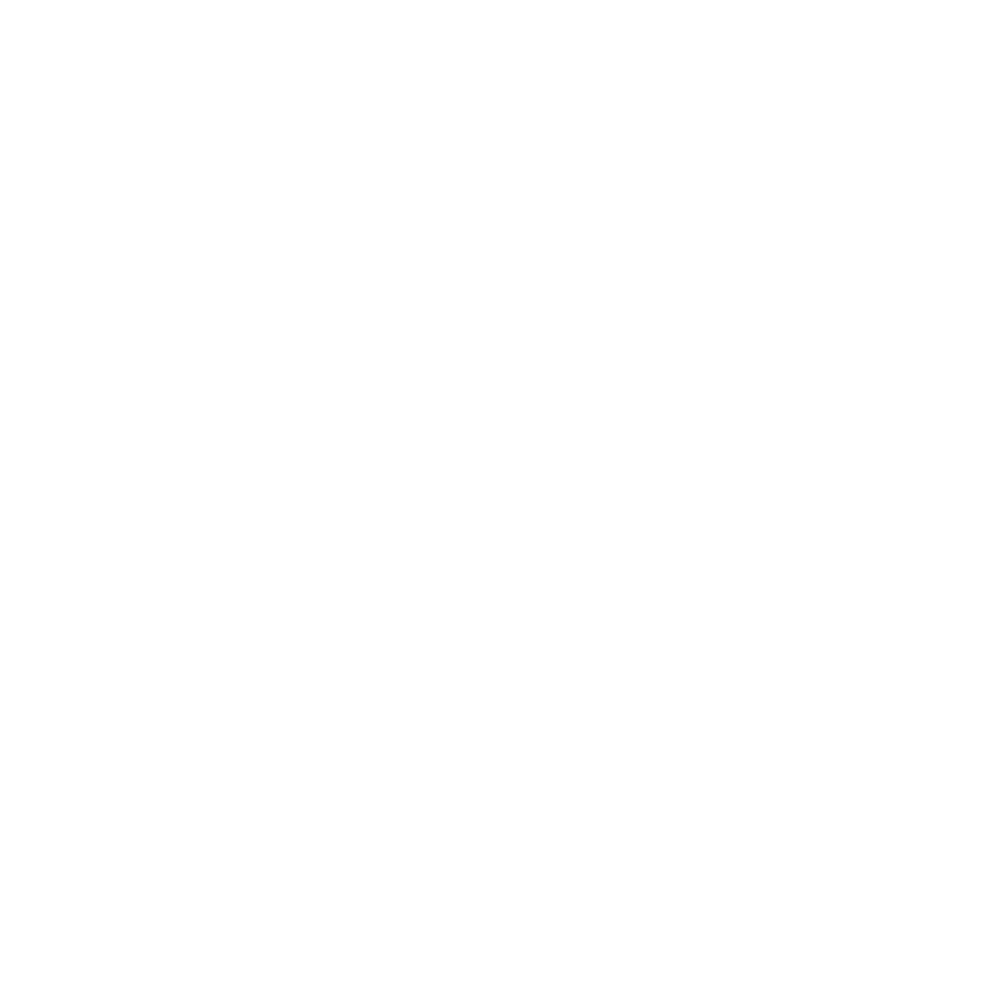}
         \end{subfigure}
         \begin{subfigure}[b]{0.4\textwidth}
             \centering
             \includegraphics[width=\linewidth]{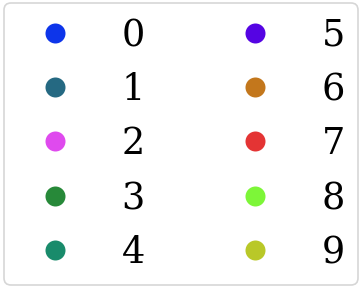}
         \end{subfigure}

     \begin{subfigure}[b]{0.2\textwidth}
         \centering
         \includegraphics[width=\linewidth]{media/empty/model.png}
     \end{subfigure}
         
     \end{subfigure}

     \begin{subfigure}[b]{0.13\textwidth}
         \centering
         \includegraphics[width=\linewidth]{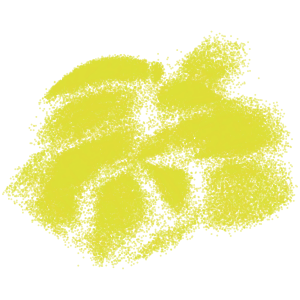}
         \caption{}
         \label{fig:determinants-mnist-geomreg}
     \end{subfigure}
     \begin{subfigure}[b]{0.13\textwidth}
         \centering
         \includegraphics[width=\linewidth]{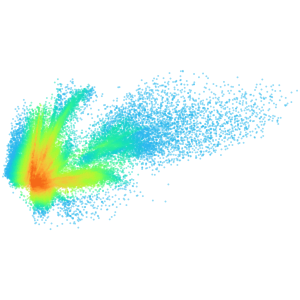}
         \caption{}
         \label{fig:determinant-mnist-vanilla}
     \end{subfigure}
     \begin{subfigure}[b]{0.13\textwidth}
         \centering
         \includegraphics[width=\linewidth]{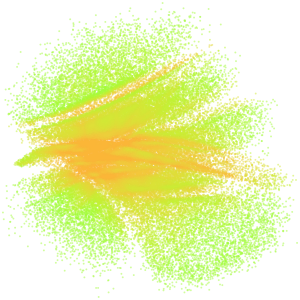}
         \caption{}
     \end{subfigure}
     \begin{subfigure}[b]{0.13\textwidth}
         \centering
         \includegraphics[width=\linewidth]{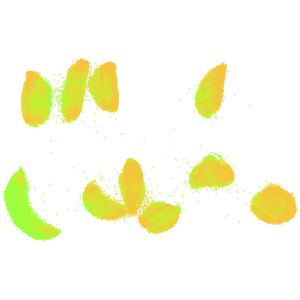}
         \caption{}
     \end{subfigure}
     \begin{subfigure}[b]{0.13\textwidth}
         \centering
         \includegraphics[width=\linewidth]{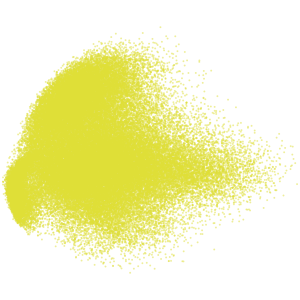}
         \caption{}
     \end{subfigure}
     \begin{subfigure}[b]{0.26\textwidth}
        \begin{subfigure}{\textwidth}
         \centering
         \includegraphics[width=\linewidth]{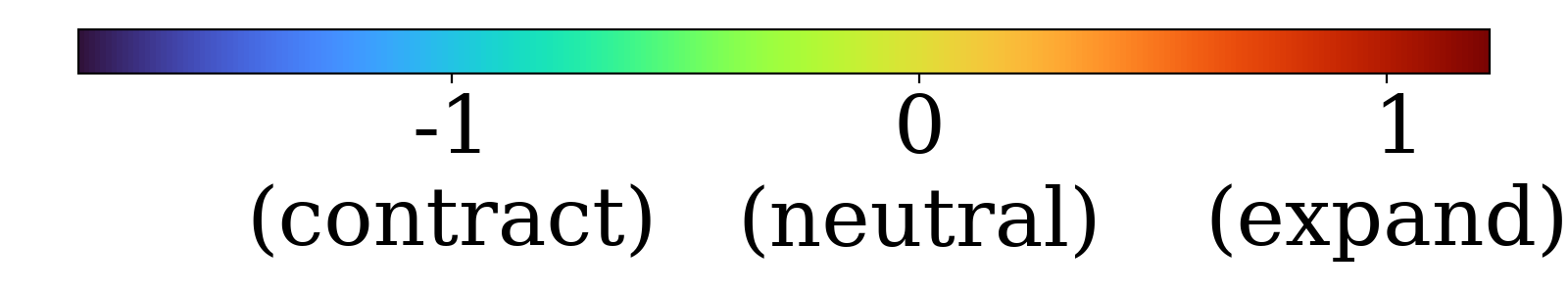}
         \end{subfigure}

        \begin{subfigure}{0.2\textwidth}
         \centering
         \includegraphics[width=\linewidth]{media/empty/model.png}
         \end{subfigure}
     \end{subfigure}

    \caption{Comparing different embedding methods on MNIST. From left to right we consider the geometric, the vanilla, the topological autoencoder, and the UMAP autoencoder followed by PCA, $t$-SNE, and UMAP. From top to bottom we show the embedding, the indicatrices and the generalized Jacobian determinant. We show our diagnostics only for those methods with a decoder.}
    \label{fig:main-evaluation}
\end{center}
\vskip -0.2in
\end{figure*}

\textbf{Qualitative Evaluation} We evaluate the geometric autoencoder as well as suitable baselines using indicatrices and the generalized Jacobian determinant. For the latter we create a heatmap plot of the logarithm of the generalized Jacobian determinant in units of their mean, and subtract $1$ from the result in order to center the scale. All values outside of the $5\%$ quantiles are collapsed to the extreme values inside the quantiles. Our results are shown in Figures~\ref{fig:main-evaluation},~\ref{fig:determinants},~\ref{fig:indicatrices}.


\textbf{Quantitative Evaluation} We evaluate all models' embeddings with metrics from~\citet{topoae} and~\citet{tsne-single-cell}.
Our local metrics are \textit{kNN}, \textit{Trust} and $\mathit{KL_{0.1}}$. The \textit{kNN} metric calculates the kNN recall from embedding to input.
\textit{Trust} is a metric based on the $k$ nearest neighbor rankings, and $\mathit{KL_{0.1}}$ measures the Kullback-Leibler divergence based on density estimates in input- and latent-space with length-scale $0.1$. Our global metrics are $\mathit{KL_{100}}$, \textit{Stress}, and \textit{Spear}.
The $\mathit{KL_{100}}$ metric considers a density estimate on a more global scale.
The \textit{Stress} metric is the loss of multidimensional scaling, and \textit{Spear} calculates the Spearman coefficient between distances in input and embedding. For a more detailed discussion of the metrics, see Appendix~\ref{sec:appendix-metrics}.

We train on and visualize all data, and evaluate the metrics on a $10\%$ random subset for speed reasons.
To aggregate results of different metrics and datasets, we rank the models for each metric and average over all datasets. Averaging these aggregated ranks over all metrics gives our final metric $\langle \text{Rank} \rangle$. The results can be found in Tables~\ref{table:aggregated-metrics} and~\ref{table:metrics}.

\begin{table*}[t]
\caption{Quantitative evaluation of our method. We rank each method for a given metric and calculate the mean over all datasets. The \mbox{$\langle$\textsc{rank}$\rangle$} is the average over metrics. Bold and underlined indicates first, bold second place.}
\label{table:aggregated-metrics}
\vskip 0.15in
\begin{center}
\begin{small}
\begin{sc}
\begin{tabular}{llllllll}
\toprule
 & \multicolumn{3}{c}{Local} & \multicolumn{3}{c}{Global} & \\
 \midrule
{} &               $\dkl_{0.1}$ &                        kNN &               Trust &                Stress &               $\dkl_{100}$ &                      Spear & $\langle \text{rank} \rangle$ \\
\midrule
Geom AE (ours)         &   \underline{\textbf{2.6}} &                        3.4 &               \textbf{2.2} &                        3.4 &   \underline{\textbf{2.2}} &                        3.4 &    \underline{\textbf{2.9}} \\
Vanilla AE        &                        5.4 &                        5.4 &                        4.4 &                        6.2 &                        4.8 &                        5.0 &                           5.2 \\
Topo AE        &               \textbf{2.8} &                        4.8 &                        4.2 &                        4.8 &               \textbf{2.2} &               \textbf{1.8} &                \textbf{3.4} \\
UMAP AE &                        4.4 &   \underline{\textbf{1.6}} &   \underline{\textbf{1.8}} &               \textbf{2.6} &                        6.0 &                        5.0 &                         3.6 \\
UMAP           &                        5.2 &                        3.4 &                        4.0 &   \underline{\textbf{1.6}} &                        5.6 &                        4.2 &                           4.0 \\
$t$-SNE           &                        4.0 &               \textbf{2.4} &                        4.4 &                        6.8 &                        3.8 &                        7.0 &                         4.7 \\
PCA            &                        3.6 &                        7.0 &                        7.0 &                        2.6 &                        3.4 &   \underline{\textbf{1.6}} &                           4.2 \\
\bottomrule
\end{tabular}
\end{sc}
\end{small}
\end{center}
\vskip -0.1in
\end{table*}

\section{Results}
\label{results}
\textbf{Qualitative Results}
In the following paragraphs we evaluate the performance of the autoencoders based on our proposed diagnostics, as well as the embeddings themselves. We computed them for all datasets considered, see Figures~\ref{fig:embeddings}--\ref{fig:indicatrices}. In the main paper, we illustrate our findings with the MNIST dataset for which we depict the embeddings, determinant heat maps and indicatrices in Figure~\ref{fig:main-evaluation}.

The geometric autoencoder produces balanced embeddings for all the datasets considered, see first column of Figure~\ref{fig:embeddings}. Its class separation is better than that of the vanilla autoencoder and PCA, slightly better than that of the topological autoencoder, but worse than $t$-SNE's and UMAP's.

While the geometric autoencoder produces visually pleasing embeddings, its true strength becomes apparent when combining it with the information obtained from our diagnostics, the determinant heatmap and the indicatrices.
As we regularize the autoencoder by the generalized Jacobian determinant, it is reassuring to see that this determinant indeed varies very little for the final embedding, see Figure~\ref{fig:determinants-mnist-geomreg} and the first column of Figure~\ref{fig:determinants}. Visually, the determinant heatmap looks as uniform as that of PCA, which has constant generalized Jacobian determinant. As a result, we can trust the relative sizes in the geometric autoencoders' plots. 

The determinant heatmaps on the vanilla autoeocoders' plots (Figure~\ref{fig:determinants}, column 2) are striking and allow us to understand the corresponding embeddings much better. Those embeddings have an extremely crowded area and a few data points or classes which take up most of the embedding space in common. For instance, it appears that the embedding of the digit $2$ (pink) in Figure~\ref{fig:embeddings-mnist-vanilla} takes up roughly as much space as the rest of the embedding. This observation is similar to the distorted continent sizes in Figure~\ref{fig:one-earth}. Consulting the determinant heatmap prevents from false conclusions: The decoder massively expands the cluttered region and contracts those clusters that take up most of the embedding area. This means that the data embedded into the cluttered area takes up much more space than appears in the embedding and conversely, the embedding of the digit $2$ in Figure~\ref{fig:embeddings-mnist-vanilla} do not actually take up much more space than the other classes. Indeed, in the embedding of the geometric autoencoder, which does not distort relative sizes, the different classes of MNIST are depicted roughly equisized.

Both the topological autoencoder and the UMAP autoencoder show more variation in local contraction and repulsion than the geometric autoencoder, see Figure~\ref{fig:determinants} columns 3 and 4. For the topological autoencoder, we see that the typically densely packed center of the embedding gets expanded by the decoder. A more spread-out layout would improve the faithfulness of the embedding. For the UMAP autoencoder, it is generally the boundary of clusters that appears too contracted, indicating that the actual cluster separation is exaggerated in the UMAP plot. Similarly, the indicatrices show that the whitespace gets contracted by the decoder.

The information encoded in the indicatrices refines the interpretation of the various embeddings further. For the geometric autoencoder, indicatrices have mostly the same area, reflecting the uniform determinant heatmaps (column 1 of Figure~\ref{fig:indicatrices}). Even though not regularized for this explicitly, they are also more circular than for other methods. This demonstrates that the geometric autoencoder does little directed stretching, leading to a recognizable world map of the Earth dataset (Figure~\ref{fig:one-earth}). We measured the mean $2$-norm condition number on the MNIST dataset for the pullback metric and found that our method, after PCA, has the most isotropic indicatrices (see Table~\ref{table:isotropy}). Analyzing the positions where an indicatrix is elongated helps us to correctly understand our embeddings.
On MNIST, the indicatrix in the long protrusion of the embedding of the digit $0$ (blue) in Figure~\ref{fig:embeddings-mnist-geomreg} is elongated in the same direction as the protrusion.
This implies that the protrusion should not be as long in the real data as depicted in the embedding.

The indicatrices for the vanilla autoencoder help us understand its embedding better: Not only is the pink class in Figure~\ref{fig:embeddings-mnist-vanilla} depicted deceitfully large, but, in particular, stretched too much horizontally. Similarly, the embedding of the digit $7$ (red) is stretched vertically. Jointly, the indicatrices seem to point towards the most densely packed region of the vanilla autoencoder embeddings for all datasets (see Figure~\ref{fig:indicatrices} column 2). Investigating further, we noticed that this dense region is typically close to the origin in embedding space. We found that at the beginning of training all embedding points are clustered tightly around the origin due the initialization of the network. During training, some classes separate by ``expanding away'' from the origin, while others stay near the origin. This explains the typical ``star-shape'' of vanilla autoencoder embeddings. We include a video of the training of a vanilla autoencoder on MNIST illustrating this process in the GitHub repository. Overall, our diagnostics enabled us to unravel the peculiar appearance of vanilla autoencoder plots.

The indicatrices for PCA are perfect circles, since the decoder of PCA consists of a pair of orthonormal vectors. Thus, PCA scores perfectly in our diagnostics, but produces embeddings with the least structure. This shows that a certain level non-linearity is necessary for salient feature extraction.

Spotting artefacts in the data is a major use-case of data visualization. Indeed, the embedding of the initial version of the PBMC dataset revealed suspiciously regular structures in the data, which turned out to be a preprocessing artefact, compare Figures~\ref{fig:pbmc-artefacts},~\subref{fig:pbmc-no-artefacts}. The geometric autoencoder highlights this artefact particularly well (Figure ~\ref{fig:pbmc-disguised}). UMAP, for example, disguises it completely.

Our geometric autoencoder visualizes semantic information even on a finer level than digit class in MNIST. For instance, it separates digits $2$ with straight lower stroke from those with curved lower strokes. The vanilla autoencoder fails to depict such subtle structure successfully. See Appendix~\ref{section:appendix-semantic-info} for more details.

\textbf{Quantitative Results}
Our quantitative results are reported in Tables~\ref{table:aggregated-metrics} and~\ref{table:metrics}. We find that the geometric regularizer influences the reconstruction loss slightly less than the topological autoencoder (see Table~\ref{table:metrics}). It furthermore has competitive reconstruction loss compared to the vanilla autoencoder, which shows that our regularizer does not lead to a major impairment of the reconstruction. The geometric autoencoder beats the vanilla baseline in all metrics except for the reconstruction loss. It furthermore achieves top rank in the $\mathit{KL_{\sigma}}$ metrics in both the local and more global setting, striking a good compromise between the preservation of local and global structure.
Overall, the geometric autoencoder balances the demands of the different metrics best as it achieves top aggregated rank. Its closest competitors are the topological autoencoder and the UMAP autoencoder, underlining the power of autoencoders for visualization when properly regularized.

\section{Limitations}
\label{limitations}

For the image of the decoder $D$ to be a manifold with a single chart, we require $D$ to be a smooth embedding. Choosing ELU activations ensures that the decoder is continuously differentiable. Since our regularizer penalizes the Jacobian for having zero determinant, the inverse function theorem gives us locally continuously differentiable invertibility.
Not fulfilling these assumptions rigorously impedes neither our regularization nor our visualization. We could just not call the decoder's image a manifold.

When defining the pullback metric, we additionally need to assume that the decoder is an immersion. If this was not the case and the differential failed to be injective at an embedding point $p$, then the metric tensor at $p$ would not be positive definite. Our diagnostics would detect this in the form of an infinitely flat indicatrix at $p$. However, our regularization loss would become infinite, hence mitigating the problem in practice.

\section{Discussion and Conclusion}
Low-dimensional visualization is key for understanding high-dimensional datasets. An embedding should capture the most salient features, while representing the dataset faithfully. Thus, our contribution consists of two components. We provide insightful diagnostics that allow identifying distortions in an embedding, as well as a novel regularizer mitigating them. The resulting embedding is more faithful when it comes to relative sizes and shapes.

Our geometric regularizer is fairly simple: It minimizes how much local expansion varies. On a range of datasets, including image and single-cell data, we used our diagnostics to show that the geometric autoencoder produces visualizations with homogeneous expansion leading to a good resolution in all parts of the embedding.

We furthermore showed that the parametric version of UMAP, when combined with an autoencoder, creates clustered and separated embeddings by contracting the dataset rather inhomogeneously, especially at the border of clusters.

\textbf{Future Work}
Equipping latent space with a metric allows for a variety of geometric diagnostics of the decoder different from our proposed indicatrices and the generalized Jacobian determinant. For example, it allows us to measure the decoder's curvature or to perform parallel transport, which is a way of moving coordinate systems ``parallel'' along a curved manifold. We imagine sampling an orthogonal coordinate system at an arbitrary point in latent space and then parallel transporting it along geodesics. The result would be a ``curved'' grid on the latent space that captures the intrinsic geometry of the high-dimensional dataset. We believe that such a latitude-longitude-like grid would be highly informative, and hope to overcome the numerical challenges encountered with existing parallel transport implementations~\cite{schilds-ladder, geomstats}.

\textbf{Acknowledgements}
This work is supported by the Deutsche Forschungsgemeinschaft (DFG, German Research Foundation) under Germany's Excellence Strategy EXC 2181/1 - 390900948 (the Heidelberg STRUCTURES Excellence Cluster) as well as by ``Informatics for Life'', funded by the Klaus Tschira Foundation.

\bibliography{references}
\bibliographystyle{icml2023}

\clearpage
\appendix
\renewcommand{\thefigure}{S\arabic{figure}}
\setcounter{figure}{0}

\renewcommand{\thetable}{S\arabic{table}}
\setcounter{table}{0}

\onecolumn
\section{Extended Figures}
\label{sec:appendix-more-datasets}
\begin{figure}[ht]
\vskip 0.2in
\centering
\begin{center}
    \begin{subfigure}{\textwidth}
        \centering
         \begin{subfigure}[b]{0.05\textwidth}
            \centering
                 \begin{tikzpicture}[font=\tiny]
                    \node(1)[rotate=90] {};
                 \end{tikzpicture}
         \end{subfigure}
         \begin{subfigure}[b]{0.12\textwidth}
            \centering
                 \begin{tikzpicture}[font=\tiny]
                    \node[minimum width=\textwidth] {Geom AE};
                 \end{tikzpicture}
         \end{subfigure}
         \hspace{.5mm}
         \begin{subfigure}[b]{0.12\textwidth}
            \centering
                 \begin{tikzpicture}[font=\tiny]
                    \node[minimum width=\textwidth] {Vanilla AE};
                 \end{tikzpicture}
         \end{subfigure}
         \hspace{.5mm}
         \begin{subfigure}[b]{0.12\textwidth}
            \centering
                 \begin{tikzpicture}[font=\tiny]
                    \node[minimum width=\textwidth] {Topo AE};
                 \end{tikzpicture}
         \end{subfigure}
         \hspace{.5mm}
         \begin{subfigure}[b]{0.12\textwidth}
            \centering
             \begin{tikzpicture}[font=\tiny]
                \node[minimum width=\textwidth] {UMAP AE};
             \end{tikzpicture}
         \end{subfigure}
         \hspace{.5mm}
         \begin{subfigure}[b]{0.12\textwidth}
            \centering
                 \begin{tikzpicture}[font=\tiny]
                    \node[minimum width=\textwidth] {PCA};
                 \end{tikzpicture}
         \end{subfigure}
         \hspace{.5mm}
         \begin{subfigure}[b]{0.12\textwidth}
            \centering
             \begin{tikzpicture}[font=\tiny]
                \node[minimum width=\textwidth] {$t$-SNE};
             \end{tikzpicture}
         \end{subfigure}
         \hspace{.5mm}
         \begin{subfigure}[b]{0.12\textwidth}
            \centering
             \begin{tikzpicture}[font=\tiny]
                \node[minimum width=\textwidth] {UMAP};
             \end{tikzpicture}
         \end{subfigure}

         \begin{subfigure}[b]{0.05\textwidth}
            \centering
                 \begin{tikzpicture}[font=\tiny]
                    \node(1)[minimum width=2.4\textwidth, rotate=90] {MNIST};
                 \end{tikzpicture}
         \end{subfigure}
         \begin{subfigure}[b]{0.12\textwidth}
             \centering
             \includegraphics[width=\linewidth]{media/mnist/geomreg/latents.png}
         \end{subfigure}
         \hspace{.5mm}
         \begin{subfigure}[b]{0.12\textwidth}
             \centering
             \includegraphics[width=\linewidth]{media/mnist/vanilla/latents.png}
         \end{subfigure}
         \hspace{.5mm}
         \begin{subfigure}[b]{0.12\textwidth}
             \centering
             \includegraphics[width=\linewidth]{media/mnist/toporeg/latents.png}
         \end{subfigure}
         \hspace{.5mm}
         \begin{subfigure}[b]{0.12\textwidth}
             \centering
             \includegraphics[width=\linewidth]{media/mnist/pumap/latents.png}
         \end{subfigure}
         \hspace{.5mm}
         \begin{subfigure}[b]{0.12\textwidth}
             \centering
             \includegraphics[width=\linewidth]{media/mnist/pca/latents.png}
         \end{subfigure}
         \hspace{.5mm}
         \begin{subfigure}[b]{0.12\textwidth}
             \centering
             \includegraphics[width=\linewidth]{media/mnist/tsne/latents.png}
         \end{subfigure}
         \hspace{.5mm}
         \begin{subfigure}[b]{0.12\textwidth}
             \centering
             \includegraphics[width=\linewidth]{media/mnist/umap/latents.png}
         \end{subfigure}
         
        \vspace{2mm}
        
         \begin{subfigure}[b]{0.05\textwidth}
            \centering
                 \begin{tikzpicture}[font=\tiny]
                    \node(1)[minimum width=2.4\textwidth, rotate=90] {FashionMNIST};
                 \end{tikzpicture}
         \end{subfigure}
        \begin{subfigure}[b]{0.12\textwidth}
             \centering
             \includegraphics[width=\linewidth]{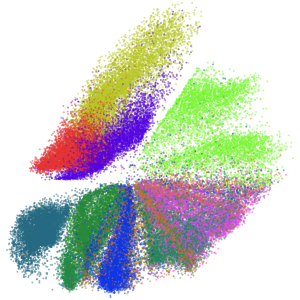}
         \end{subfigure}
         \hspace{.5mm}
         \begin{subfigure}[b]{0.12\textwidth}
             \centering
             \includegraphics[width=\linewidth]{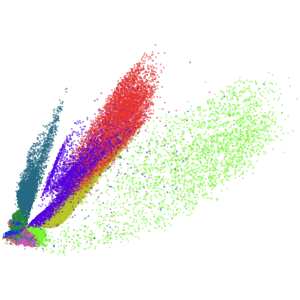}
         \end{subfigure}
         \hspace{.5mm}
         \begin{subfigure}[b]{0.12\textwidth}
             \centering
             \includegraphics[width=\linewidth]{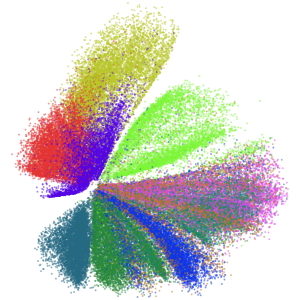}
         \end{subfigure}
         \hspace{.5mm}
         \begin{subfigure}[b]{0.12\textwidth}
             \centering
             \includegraphics[width=\linewidth]{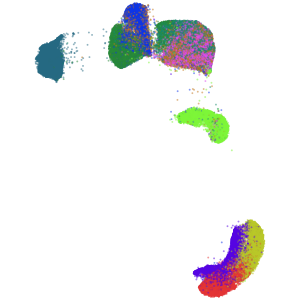}
         \end{subfigure}
         \hspace{.5mm}
         \begin{subfigure}[b]{0.12\textwidth}
             \centering
             \includegraphics[width=\linewidth]{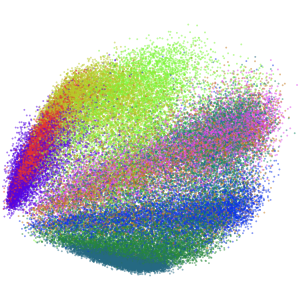}
         \end{subfigure}
         \hspace{.5mm}
         \begin{subfigure}[b]{0.12\textwidth}
             \centering
             \includegraphics[width=\linewidth]{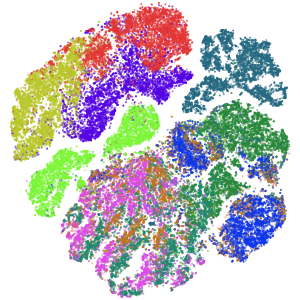}
         \end{subfigure}
         \hspace{.5mm}
         \begin{subfigure}[b]{0.12\textwidth}
             \centering
             \includegraphics[width=\linewidth]{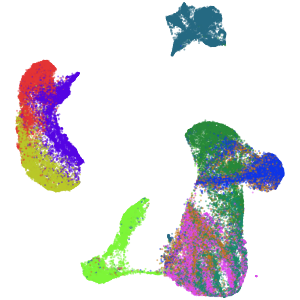}
         \end{subfigure}
         
        \vspace{2mm}
        
         \begin{subfigure}[b]{0.05\textwidth}
            \centering
                 \begin{tikzpicture}[font=\tiny]
                    \node(1)[minimum width=2.4\textwidth, rotate=90] {Zilionis};
                 \end{tikzpicture}
         \end{subfigure}
        \begin{subfigure}[b]{0.12\textwidth}
             \centering
             \includegraphics[width=\linewidth]{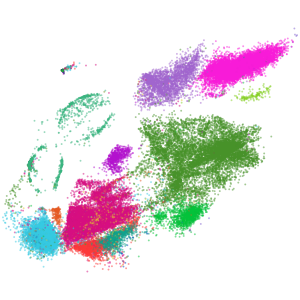}
         \end{subfigure}
         \hspace{.5mm}
         \begin{subfigure}[b]{0.12\textwidth}
             \centering
             \includegraphics[width=\linewidth]{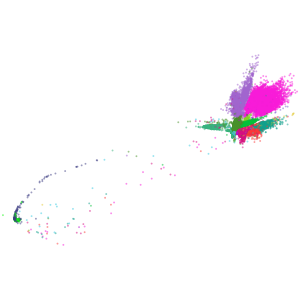}
         \end{subfigure}
         \hspace{.5mm}
         \begin{subfigure}[b]{0.12\textwidth}
             \centering
             \includegraphics[width=\linewidth]{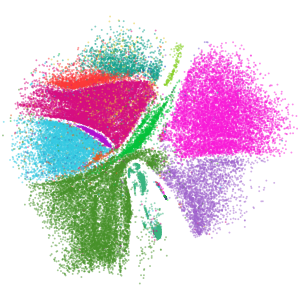}
         \end{subfigure}
         \hspace{.5mm}
         \begin{subfigure}[b]{0.12\textwidth}
             \centering
             \includegraphics[width=\linewidth]{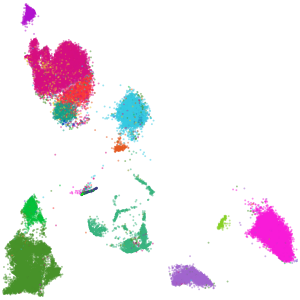}
         \end{subfigure}
         \hspace{.5mm}
         \begin{subfigure}[b]{0.12\textwidth}
             \centering
             \includegraphics[width=\linewidth]{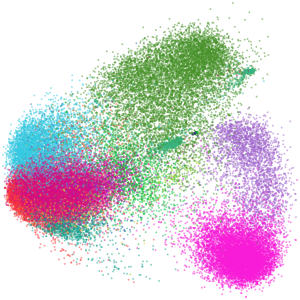}
         \end{subfigure}
         \hspace{.5mm}
         \begin{subfigure}[b]{0.12\textwidth}
             \centering
             \includegraphics[width=\linewidth]{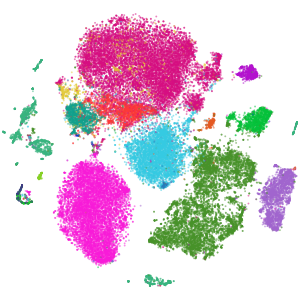}
         \end{subfigure}
         \hspace{.5mm}
         \begin{subfigure}[b]{0.12\textwidth}
             \centering
             \includegraphics[width=\linewidth]{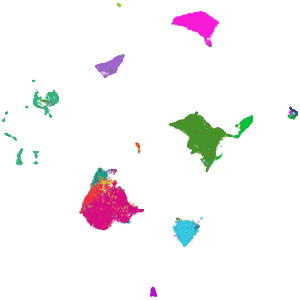}
         \end{subfigure}
    
        \vspace{2mm}
        
         \begin{subfigure}[b]{0.05\textwidth}
            \centering
                 \begin{tikzpicture}[font=\tiny]
                    \node(1)[minimum width=2.4\textwidth, rotate=90] {PBMC};
                 \end{tikzpicture}
         \end{subfigure}
        \begin{subfigure}[b]{0.12\textwidth}
             \centering
             \includegraphics[width=\linewidth]{media/pbmc/geomreg/latents.png}
         \end{subfigure}
         \hspace{.5mm}
         \begin{subfigure}[b]{0.12\textwidth}
             \centering
             \includegraphics[width=\linewidth]{media/pbmc/vanilla/latents.png}
         \end{subfigure}
         \hspace{.5mm}
         \begin{subfigure}[b]{0.12\textwidth}
             \centering
             \includegraphics[width=\linewidth]{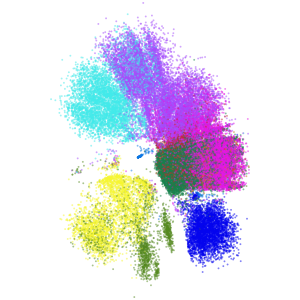}
         \end{subfigure}
         \hspace{.5mm}
         \begin{subfigure}[b]{0.12\textwidth}
             \centering
             \includegraphics[width=\linewidth]{media/pbmc/pumap/latents.png}
         \end{subfigure}
         \hspace{.5mm}
         \begin{subfigure}[b]{0.12\textwidth}
             \centering
             \includegraphics[width=\linewidth]{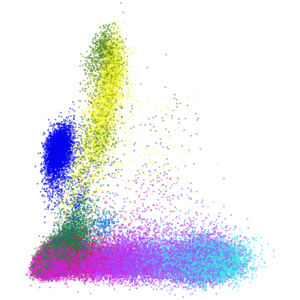}
         \end{subfigure}
         \hspace{.5mm}
         \begin{subfigure}[b]{0.12\textwidth}
             \centering
             \includegraphics[width=\linewidth]{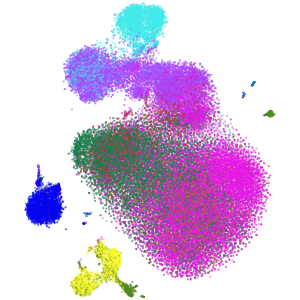}
         \end{subfigure}
         \hspace{.5mm}
         \begin{subfigure}[b]{0.12\textwidth}
             \centering
             \includegraphics[width=\linewidth]{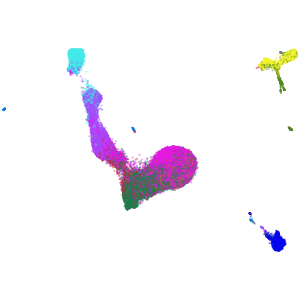}
         \end{subfigure}
    
        \vspace{2mm}
        
         \begin{subfigure}[b]{0.05\textwidth}
            \centering
                 \begin{tikzpicture}[font=\tiny]
                    \node(1)[minimum width=2.4\textwidth, rotate=90] {CElegans};
                 \end{tikzpicture}
         \end{subfigure}
        \begin{subfigure}[b]{0.12\textwidth}
             \centering
             \includegraphics[width=\linewidth]{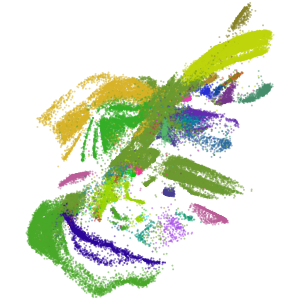}
         \end{subfigure}
         \hspace{.5mm}
         \begin{subfigure}[b]{0.12\textwidth}
             \centering
             \includegraphics[width=\linewidth]{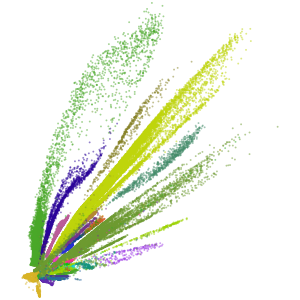}
         \end{subfigure}
         \hspace{.5mm}
         \begin{subfigure}[b]{0.12\textwidth}
             \centering
             \includegraphics[width=\linewidth]{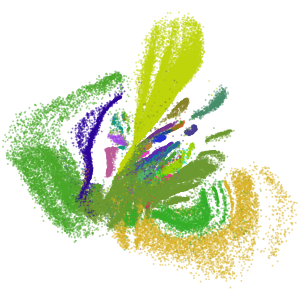}
         \end{subfigure}
         \hspace{.5mm}
         \begin{subfigure}[b]{0.12\textwidth}
             \centering
             \includegraphics[width=\linewidth]{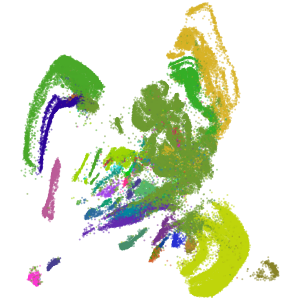}
         \end{subfigure}
         \hspace{.5mm}
         \begin{subfigure}[b]{0.12\textwidth}
             \centering
             \includegraphics[width=\linewidth]{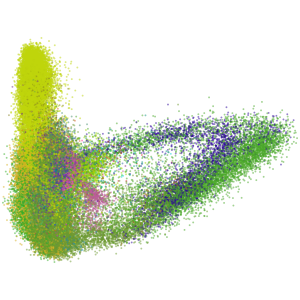}
         \end{subfigure}
         \hspace{.5mm}
         \begin{subfigure}[b]{0.12\textwidth}
             \centering
             \includegraphics[width=\linewidth]{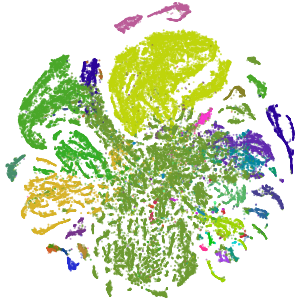}
         \end{subfigure}
         \hspace{.5mm}
         \begin{subfigure}[b]{0.12\textwidth}
             \centering
             \includegraphics[width=\linewidth]{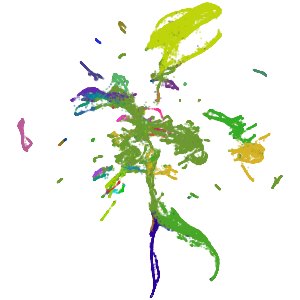}
         \end{subfigure}
    \end{subfigure}
    \caption{Embeddings of all datasets created with all models.}
    \label{fig:embeddings}
\end{center}
\vskip -0.2in
\end{figure}

\begin{figure*}
\vskip 0.2in
\begin{center}
    \begin{subfigure}{.8\textwidth}
        \centering
         \begin{subfigure}[b]{0.05\textwidth}
            \centering
                 \begin{tikzpicture}[font=\tiny]
                    \node(1)[rotate=90] {};
                 \end{tikzpicture}
         \end{subfigure}
         \begin{subfigure}[b]{0.15\textwidth}
            \centering
                 \begin{tikzpicture}[font=\tiny]
                    \node[minimum width=\textwidth] {Geom AE};
                 \end{tikzpicture}
        \end{subfigure}
        \hspace{.5mm}
         \begin{subfigure}[b]{0.15\textwidth}
            \centering
                 \begin{tikzpicture}[font=\tiny]
                    \node[minimum width=\textwidth] {Vanilla AE};
                 \end{tikzpicture}
         \end{subfigure}
         \hspace{.5mm}
         \begin{subfigure}[b]{0.15\textwidth}
            \centering
                 \begin{tikzpicture}[font=\tiny]
                    \node[minimum width=\textwidth] {Topo AE};
                 \end{tikzpicture}
         \end{subfigure}
         \hspace{.5mm}
         \begin{subfigure}[b]{0.15\textwidth}
            \centering
             \begin{tikzpicture}[font=\tiny]
                \node[minimum width=\textwidth] {UMAP AE};
             \end{tikzpicture}
         \end{subfigure}
         \hspace{.5mm}
         \begin{subfigure}[b]{0.15\textwidth}
            \centering
                 \begin{tikzpicture}[font=\tiny]
                    \node[minimum width=\textwidth] {PCA};
                 \end{tikzpicture}
         \end{subfigure}

         \begin{subfigure}[b]{0.05\textwidth}
            \centering
                 \begin{tikzpicture}[font=\tiny]
                    \node(1)[minimum width=3\textwidth, rotate=90] {MNIST};
                 \end{tikzpicture}
         \end{subfigure}
         \begin{subfigure}[b]{0.15\textwidth}
             \centering
             \includegraphics[width=\linewidth]{media/mnist/geomreg/det.png}
         \end{subfigure}
        \hspace{.5mm}
         \begin{subfigure}[b]{0.15\textwidth}
             \centering
             \includegraphics[width=\linewidth]{media/mnist/vanilla/det.png}
         \end{subfigure}
         \hspace{.5mm}
         \begin{subfigure}[b]{0.15\textwidth}
             \centering
             \includegraphics[width=\linewidth]{media/mnist/toporeg/det.png}
         \end{subfigure}
         \hspace{.5mm}
         \begin{subfigure}[b]{0.15\textwidth}
             \centering
             \includegraphics[width=\linewidth]{media/mnist/pumap/det.png}
         \end{subfigure}
         \hspace{.5mm}
         \begin{subfigure}[b]{0.15\textwidth}
             \centering
             \includegraphics[width=\linewidth]{media/mnist/pca/det.png}
         \end{subfigure}
        
        \vspace{2mm}
        
         \begin{subfigure}[b]{0.05\textwidth}
            \centering
                 \begin{tikzpicture}[font=\tiny]
                    \node(1)[minimum width=3\textwidth, rotate=90] {FashionMNIST};
                 \end{tikzpicture}
         \end{subfigure}
        \begin{subfigure}[b]{0.15\textwidth}
             \centering
             \includegraphics[width=\linewidth]{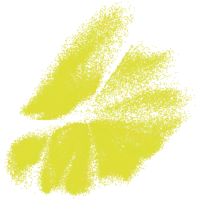}
         \end{subfigure}
         \hspace{.5mm}
         \begin{subfigure}[b]{0.15\textwidth}
             \centering
             \includegraphics[width=\linewidth]{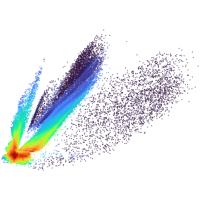}
         \end{subfigure}
         \hspace{.5mm}
         \begin{subfigure}[b]{0.15\textwidth}
             \centering
             \includegraphics[width=\linewidth]{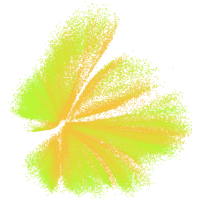}
         \end{subfigure}
         \hspace{.5mm}
         \begin{subfigure}[b]{0.15\textwidth}
             \centering
             \includegraphics[width=\linewidth]{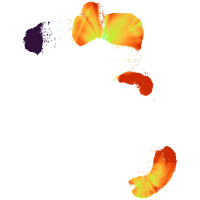}
         \end{subfigure}
         \hspace{.5mm}
         \begin{subfigure}[b]{0.15\textwidth}
             \centering
             \includegraphics[width=\linewidth]{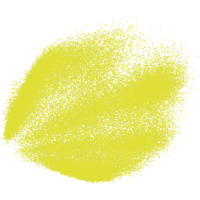}
         \end{subfigure}
        
         \vspace{2mm}
        
         \begin{subfigure}[b]{0.05\textwidth}
            \centering
                 \begin{tikzpicture}[font=\tiny]
                    \node(1)[minimum width=3\textwidth, rotate=90] {Zilionis};
                 \end{tikzpicture}
         \end{subfigure}
        \begin{subfigure}[b]{0.15\textwidth}
             \centering
             \includegraphics[width=\linewidth]{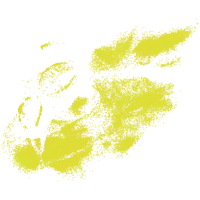}
         \end{subfigure}
         \hspace{.5mm}
         \begin{subfigure}[b]{0.15\textwidth}
             \centering
             \includegraphics[width=\linewidth]{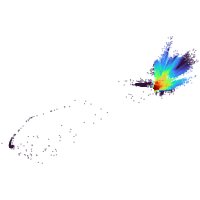}
         \end{subfigure}
         \hspace{.5mm}
         \begin{subfigure}[b]{0.15\textwidth}
             \centering
             \includegraphics[width=\linewidth]{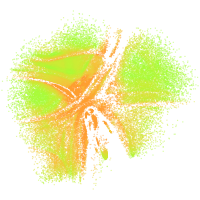}
         \end{subfigure}
         \hspace{.5mm}
         \begin{subfigure}[b]{0.15\textwidth}
             \centering
             \includegraphics[width=\linewidth]{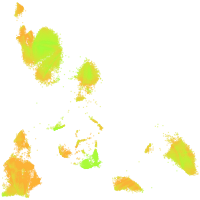}
         \end{subfigure}
         \hspace{.5mm}
         \begin{subfigure}[b]{0.15\textwidth}
             \centering
             \includegraphics[width=\linewidth]{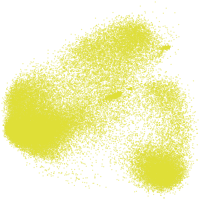}
         \end{subfigure}
        
         \vspace{2mm}
        
         \begin{subfigure}[b]{0.05\textwidth}
            \centering
                 \begin{tikzpicture}[font=\tiny]
                    \node(1)[minimum width=3\textwidth, rotate=90] {PBMC};
                 \end{tikzpicture}
         \end{subfigure}
        \begin{subfigure}[b]{0.15\textwidth}
             \centering
             \includegraphics[width=\linewidth]{media/pbmc/geomreg/det.png}
         \end{subfigure}
         \hspace{.5mm}
         \begin{subfigure}[b]{0.15\textwidth}
             \centering
             \includegraphics[width=\linewidth]{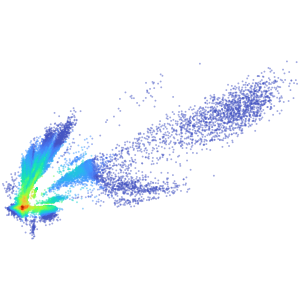}
         \end{subfigure}
         \hspace{.5mm}
         \begin{subfigure}[b]{0.15\textwidth}
             \centering
             \includegraphics[width=\linewidth]{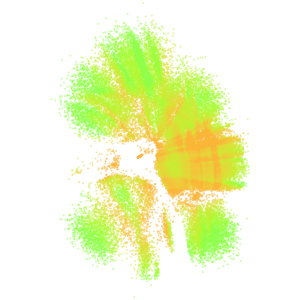}
         \end{subfigure}
         \hspace{.5mm}
         \begin{subfigure}[b]{0.15\textwidth}
             \centering
             \includegraphics[width=\linewidth]{media/pbmc/pumap/det.png}
         \end{subfigure}
         \hspace{.5mm}
         \begin{subfigure}[b]{0.15\textwidth}
             \centering
             \includegraphics[width=\linewidth]{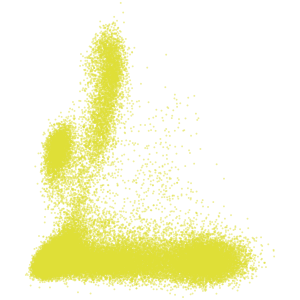}
         \end{subfigure}
        
         \vspace{2mm}
        
         \begin{subfigure}[b]{0.05\textwidth}
            \centering
                 \begin{tikzpicture}[font=\tiny]
                    \node(1)[minimum width=3\textwidth, rotate=90] {CElegans};
                 \end{tikzpicture}
         \end{subfigure}
        \begin{subfigure}[b]{0.15\textwidth}
             \centering
             \includegraphics[width=\linewidth]{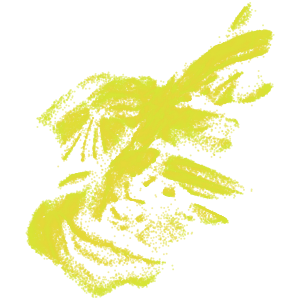}
         \end{subfigure}
         \hspace{.5mm}
         \begin{subfigure}[b]{0.15\textwidth}
             \centering
             \includegraphics[width=\linewidth]{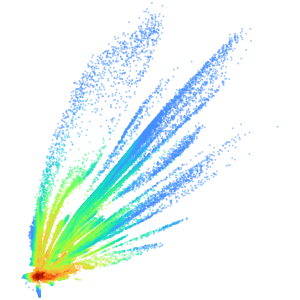}
         \end{subfigure}
         \hspace{.5mm}
         \begin{subfigure}[b]{0.15\textwidth}
             \centering
             \includegraphics[width=\linewidth]{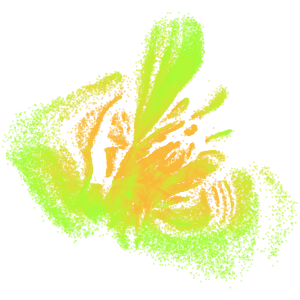}
         \end{subfigure}
         \hspace{.5mm}
         \begin{subfigure}[b]{0.15\textwidth}
             \centering
             \includegraphics[width=\linewidth]{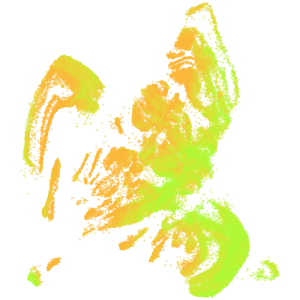}
         \end{subfigure}
         \hspace{.5mm}
         \begin{subfigure}[b]{0.15\textwidth}
             \centering
             \includegraphics[width=\linewidth]{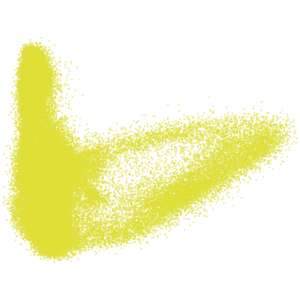}
         \end{subfigure}
         \vspace{1cm}
         \begin{subfigure}[b]{.6\textwidth}
             \centering
             \includegraphics[width=\linewidth]{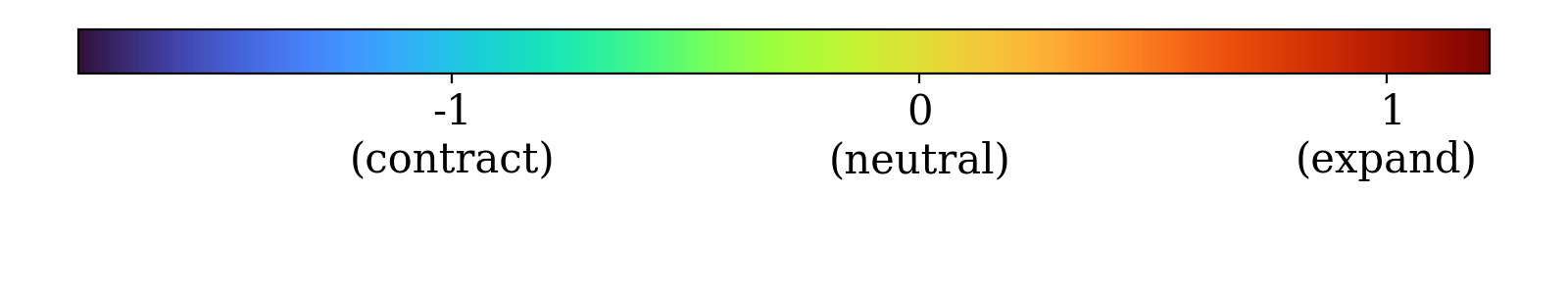}
         \end{subfigure}

    \end{subfigure}

    \caption{Determinant diagnostic for all suitable models on all datasets.}
    \label{fig:determinants}
\end{center}
\vskip -0.2in
\end{figure*}

\begin{figure*}[ht]
	\vskip 0.2in
	\begin{center}

    \begin{subfigure}{.8\textwidth}
    \centering
 
     \begin{subfigure}[b]{0.05\textwidth}
        \centering
             \begin{tikzpicture}[font=\tiny]
                \node(1)[rotate=90] {};
             \end{tikzpicture}
     \end{subfigure}
     \begin{subfigure}[b]{0.15\textwidth}
        \centering
             \begin{tikzpicture}[font=\tiny]
                \node[minimum width=\textwidth] {Geom AE};
             \end{tikzpicture}
     \end{subfigure}
     \hspace{.5mm}
     \begin{subfigure}[b]{0.15\textwidth}
        \centering
             \begin{tikzpicture}[font=\tiny]
                \node[minimum width=\textwidth] {Vanilla AE};
             \end{tikzpicture}
     \end{subfigure}
     \hspace{.5mm}
     \begin{subfigure}[b]{0.15\textwidth}
        \centering
             \begin{tikzpicture}[font=\tiny]
                \node[minimum width=\textwidth] {Topo AE};
             \end{tikzpicture}
     \end{subfigure}
     \hspace{.5mm}
     \begin{subfigure}[b]{0.15\textwidth}
        \centering
         \begin{tikzpicture}[font=\tiny]
            \node[minimum width=\textwidth] {UMAP AE};
         \end{tikzpicture}
     \end{subfigure}
     \hspace{.5mm}
     \begin{subfigure}[b]{0.15\textwidth}
        \centering
             \begin{tikzpicture}[font=\tiny]
                \node[minimum width=\textwidth] {PCA};
             \end{tikzpicture}
     \end{subfigure}

     \begin{subfigure}[b]{0.05\textwidth}
        \centering
             \begin{tikzpicture}[font=\tiny]
                \node(1)[minimum width=3\textwidth, rotate=90] {MNIST};
             \end{tikzpicture}
     \end{subfigure}
        \begin{subfigure}[b]{0.15\textwidth}
			\centering
			\includegraphics[width=\linewidth]{media/mnist/geomreg/indicatrices.png}
		\end{subfigure}
        \hspace{.5mm}
		\begin{subfigure}[b]{0.15\textwidth}
			\centering
			\includegraphics[width=\linewidth]{media/mnist/vanilla/indicatrices.png}
		\end{subfigure}
        \hspace{.5mm}
		\begin{subfigure}[b]{0.15\textwidth}
			\centering
			\includegraphics[width=\linewidth]{media/mnist/toporeg/indicatrices.png}
		\end{subfigure}
        \hspace{.5mm}
		\begin{subfigure}[b]{0.15\textwidth}
			\centering
			\includegraphics[width=\linewidth]{media/mnist/pumap/indicatrices.png}
		\end{subfigure}
        \hspace{.5mm}
		\begin{subfigure}[b]{0.15\textwidth}
			\centering
			\includegraphics[width=\linewidth]{media/mnist/pca/indicatrices.png}
		\end{subfigure}

    \vspace{2mm}

     \begin{subfigure}[b]{0.05\textwidth}
        \centering
             \begin{tikzpicture}[font=\tiny]
                \node(1)[minimum width=3\textwidth, rotate=90] {FashionMNIST};
             \end{tikzpicture}
     \end{subfigure}
		\begin{subfigure}[b]{0.15\textwidth}
			\centering
			\includegraphics[width=\linewidth]{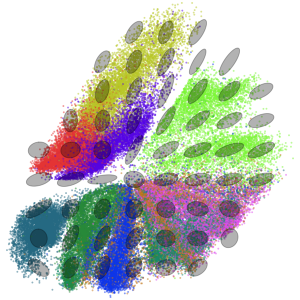}
		\end{subfigure}
  \hspace{.5mm}
		\begin{subfigure}[b]{0.15\textwidth}
			\centering
			\includegraphics[width=\linewidth]{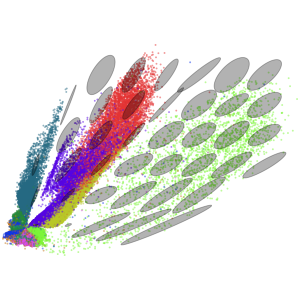}
		\end{subfigure}
  \hspace{.5mm}
		\begin{subfigure}[b]{0.15\textwidth}
			\centering
			\includegraphics[width=\linewidth]{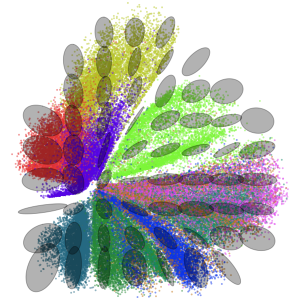}
		\end{subfigure}
  \hspace{.5mm}
		\begin{subfigure}[b]{0.15\textwidth}
			\centering
			\includegraphics[width=\linewidth]{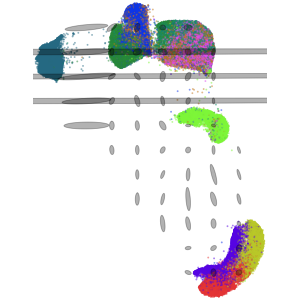}
		\end{subfigure}
  \hspace{.5mm}
		\begin{subfigure}[b]{0.15\textwidth}
			\centering
			\includegraphics[width=\linewidth]{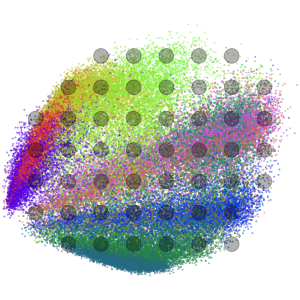}
		\end{subfigure}

  \vspace{2mm}

     \begin{subfigure}[b]{0.05\textwidth}
        \centering
             \begin{tikzpicture}[font=\tiny]
                \node(1)[minimum width=3\textwidth, rotate=90] {Zilionis};
             \end{tikzpicture}
     \end{subfigure}
		\begin{subfigure}[b]{0.15\textwidth}
			\centering
			\includegraphics[width=\linewidth]{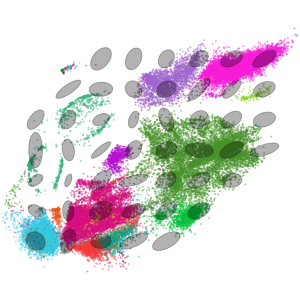}
		\end{subfigure}
  \hspace{.5mm}
		\begin{subfigure}[b]{0.15\textwidth}
			\centering
			\includegraphics[width=\linewidth]{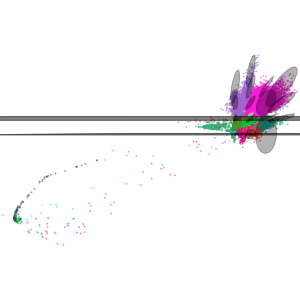}
		\end{subfigure}
  \hspace{.5mm}
		\begin{subfigure}[b]{0.15\textwidth}
			\centering
			\includegraphics[width=\linewidth]{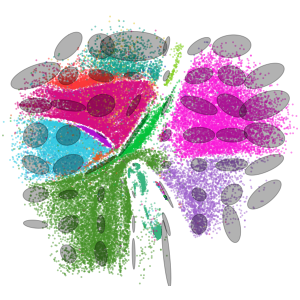}
		\end{subfigure}
  \hspace{.5mm}
		\begin{subfigure}[b]{0.15\textwidth}
			\centering
			\includegraphics[width=\linewidth]{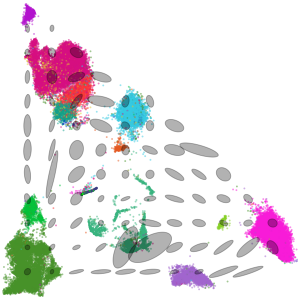}
		\end{subfigure}
  \hspace{.5mm}
		\begin{subfigure}[b]{0.15\textwidth}
			\centering
			\includegraphics[width=\linewidth]{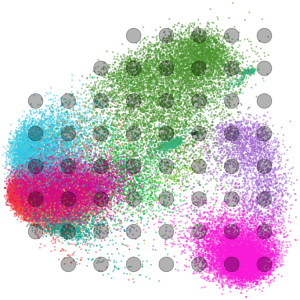}
		\end{subfigure}

    \vspace{2mm}

     \begin{subfigure}[b]{0.05\textwidth}
        \centering
             \begin{tikzpicture}[font=\tiny]
                \node(1)[minimum width=3\textwidth, rotate=90] {PBMC};
             \end{tikzpicture}
     \end{subfigure}
		\begin{subfigure}[b]{0.15\textwidth}
			\centering
			\includegraphics[width=\linewidth]{media/pbmc/geomreg/indicatrices.png}
		\end{subfigure}
  \hspace{.5mm}
		\begin{subfigure}[b]{0.15\textwidth}
			\centering
			\includegraphics[width=\linewidth]{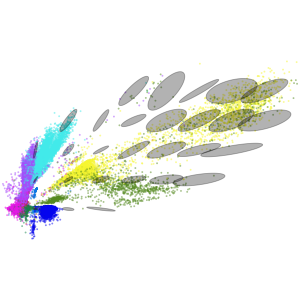}
		\end{subfigure}
  \hspace{.5mm}
		\begin{subfigure}[b]{0.15\textwidth}
			\centering
			\includegraphics[width=\linewidth]{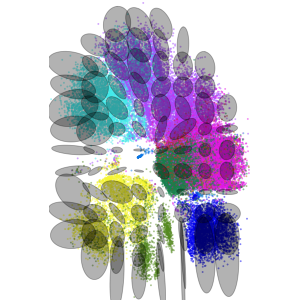}
		\end{subfigure}
  \hspace{.5mm}
		\begin{subfigure}[b]{0.15\textwidth}
			\centering
			\includegraphics[width=\linewidth]{media/pbmc/pumap/indicatrices.png}
		\end{subfigure}
  \hspace{.5mm}
		\begin{subfigure}[b]{0.15\textwidth}
			\centering
			\includegraphics[width=\linewidth]{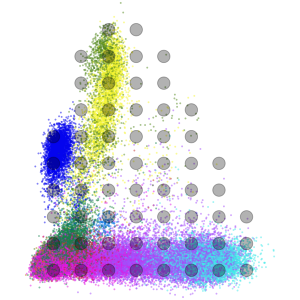}
		\end{subfigure}

  \vspace{2mm}

     \begin{subfigure}[b]{0.05\textwidth}
        \centering
             \begin{tikzpicture}[font=\tiny]
                \node(1)[minimum width=3\textwidth, rotate=90] {CElegans};
             \end{tikzpicture}
     \end{subfigure}
		\begin{subfigure}[b]{0.15\textwidth}
			\centering
			\includegraphics[width=\linewidth]{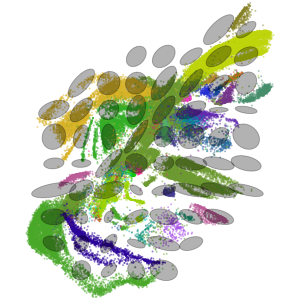}
		\end{subfigure}
  \hspace{.5mm}
		\begin{subfigure}[b]{0.15\textwidth}
			\centering
			\includegraphics[width=\linewidth]{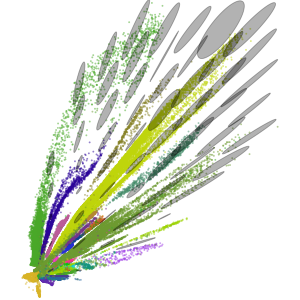}
		\end{subfigure}
  \hspace{.5mm}
		\begin{subfigure}[b]{0.15\textwidth}
			\centering
			\includegraphics[width=\linewidth]{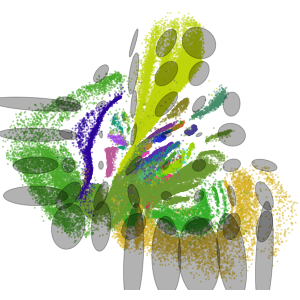}
		\end{subfigure}
  \hspace{.5mm}
		\begin{subfigure}[b]{0.15\textwidth}
			\centering
			\includegraphics[width=\linewidth]{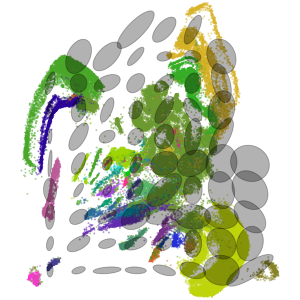}
		\end{subfigure}
  \hspace{.5mm}
		\begin{subfigure}[b]{0.15\textwidth}
			\centering
			\includegraphics[width=\linewidth]{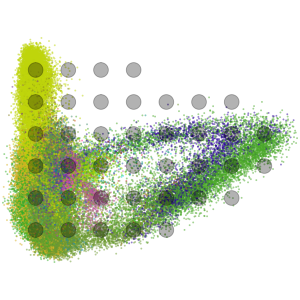}
		\end{subfigure}

    \end{subfigure}
       
		\caption{Indicatrix diagnostics for all suitable models on all datasets.}
		\label{fig:indicatrices}
	\end{center}
	\vskip -0.2in
\end{figure*}

\begin{figure*}[ht]
\vskip 0.2in
\begin{center}
    \centering
     \begin{subfigure}[b]{0.49\textwidth}
         \centering
         \includegraphics[width=\linewidth]{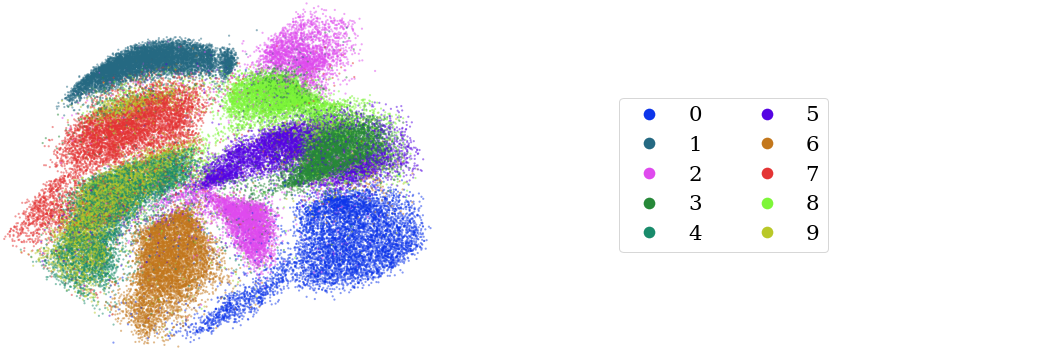}
         \caption{MNIST}
     \end{subfigure}   
     \begin{subfigure}[b]{0.49\textwidth}
         \centering
         \includegraphics[width=\linewidth]{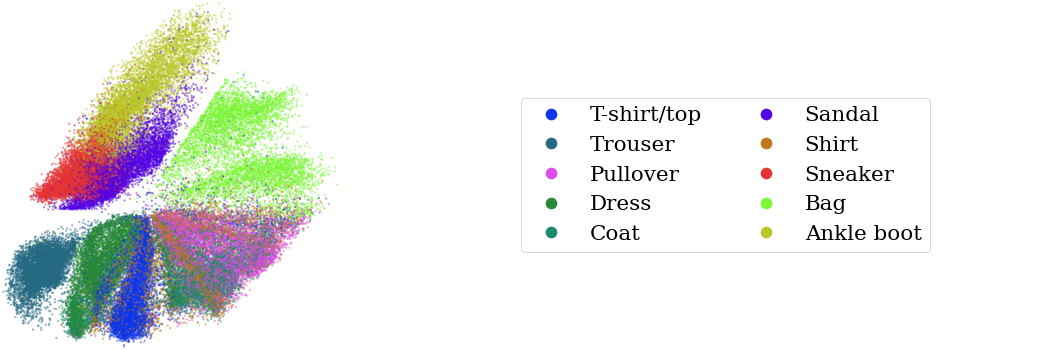}
         \caption{FashionMNIST}
     \end{subfigure}

    \begin{subfigure}[b]{0.49\textwidth}
         \centering
         \includegraphics[width=\linewidth]{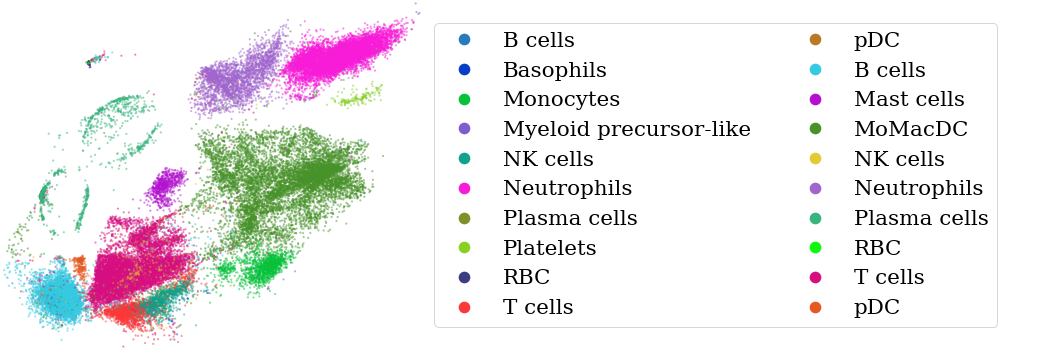}
         \caption{Zilionis}
     \end{subfigure}
    \begin{subfigure}[b]{0.49\textwidth}
         \centering
         \includegraphics[width=\linewidth]{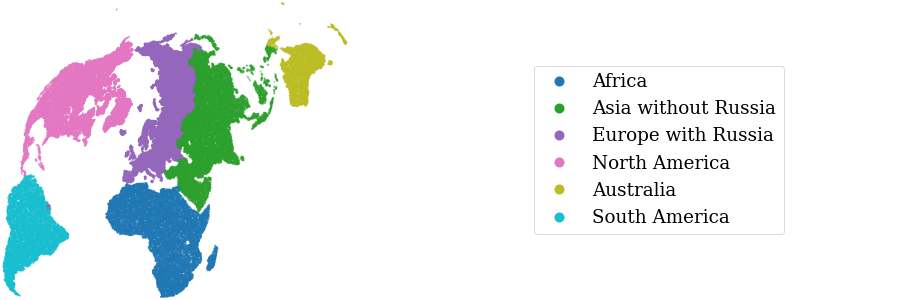}
         \caption{Earth}
     \end{subfigure}

    \begin{subfigure}[b]{\textwidth}
         \centering
         \includegraphics[width=\linewidth]{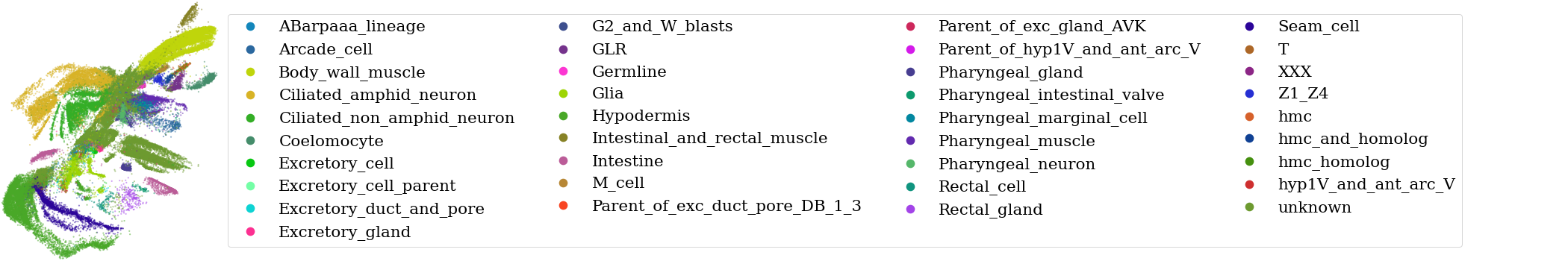}
         \caption{CElegans}
     \end{subfigure}
     
    \begin{subfigure}[b]{0.56\textwidth}
         \centering
         \includegraphics[width=\linewidth]{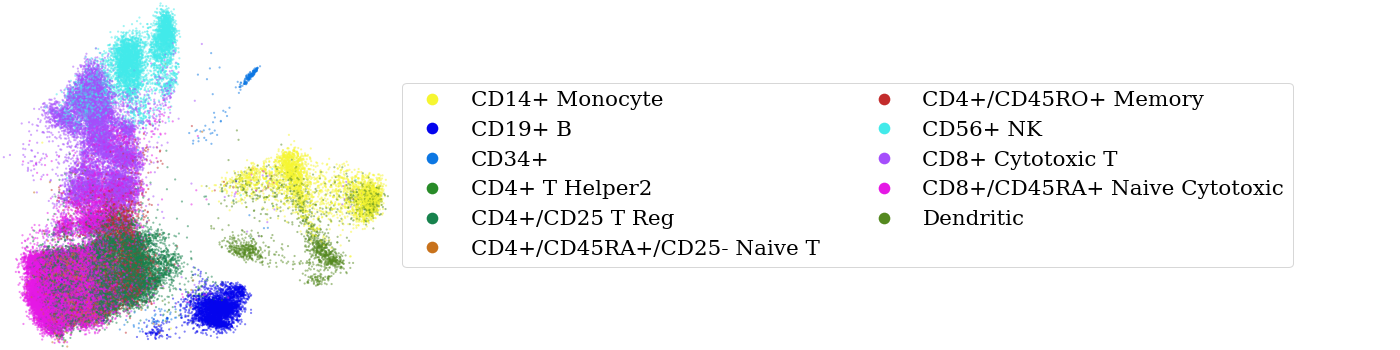}
         \caption{PBMC}
     \end{subfigure}

    \caption{Datasets with labels. Embeddings created with geometric autoencoder.}
\label{fig:labels}
\end{center}
\vskip -0.2in
\end{figure*}

In Figures~\ref{fig:embeddings},~\ref{fig:determinants} and~\ref{fig:indicatrices} we show the embeddings, determinant heatmap plots and indicatrices for all the datasets and models considered. In Figure~\ref{fig:labels} we show the labels of all datasets.

\begin{figure*}[ht]
	\vskip 0.2in
	\begin{center}

    \begin{subfigure}{.8\textwidth}
    \centering
 
     \begin{subfigure}[b]{0.05\textwidth}
        \centering
             \begin{tikzpicture}[font=\tiny]
                \node(1)[rotate=90] {};
             \end{tikzpicture}
     \end{subfigure}
     \begin{subfigure}[b]{0.35\textwidth}
        \centering
             \begin{tikzpicture}[font=\tiny]
                \node[minimum width=\textwidth] {MNIST};
             \end{tikzpicture}
     \end{subfigure}
     \hspace{.5mm}
     \begin{subfigure}[b]{0.35\textwidth}
        \centering
             \begin{tikzpicture}[font=\tiny]
                \node[minimum width=\textwidth] {FashionMNIST};
             \end{tikzpicture}
     \end{subfigure}

     \begin{subfigure}[b]{0.05\textwidth}
        \centering
             \begin{tikzpicture}[font=\tiny]
                \node(1)[minimum width=3\textwidth, rotate=90] {Vanilla AE};
             \end{tikzpicture}
     \end{subfigure}
        \begin{subfigure}[b]{0.15\textwidth}
			\centering
			\includegraphics[width=\linewidth]{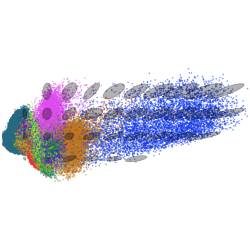}
		\end{subfigure}
        \hspace{.2mm}
        \begin{subfigure}[b]{0.15\textwidth}
			\centering
			\includegraphics[width=\linewidth]{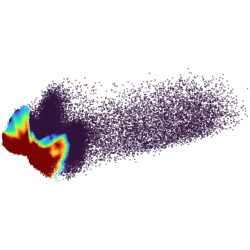}
		\end{subfigure}
        \hspace{5mm}
        \begin{subfigure}[b]{0.15\textwidth}
			\centering
			\includegraphics[width=\linewidth]{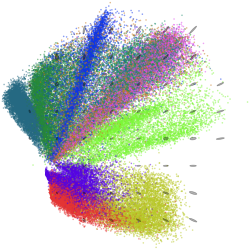}
		\end{subfigure}
        \hspace{.2mm}
        \begin{subfigure}[b]{0.15\textwidth}
			\centering
			\includegraphics[width=\linewidth]{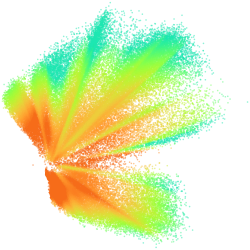}
		\end{subfigure}

    \vspace{2mm}

    \begin{subfigure}[b]{0.05\textwidth}
        \centering
             \begin{tikzpicture}[font=\tiny]
                \node(1)[minimum width=3\textwidth, rotate=90] {Geom AE};
             \end{tikzpicture}
     \end{subfigure}
        \begin{subfigure}[b]{0.15\textwidth}
			\centering
			\includegraphics[width=\linewidth]{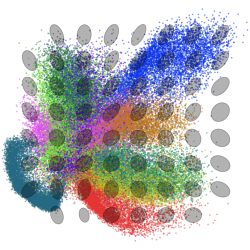}
		\end{subfigure}
        \hspace{.2mm}
        \begin{subfigure}[b]{0.15\textwidth}
			\centering
			\includegraphics[width=\linewidth]{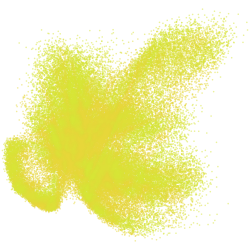}
		\end{subfigure}
      \hspace{5mm}
              \begin{subfigure}[b]{0.15\textwidth}
    			\centering
    			\includegraphics[width=\linewidth]{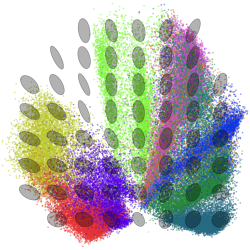}
    		\end{subfigure}
            \hspace{.2mm}
            \begin{subfigure}[b]{0.15\textwidth}
    			\centering
    			\includegraphics[width=\linewidth]{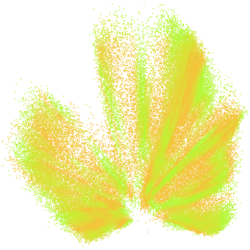}
    		\end{subfigure}  
    \end{subfigure}
		\caption{Convolutional autoencoders trained and evaluated on MNIST and FashionMNIST.}
		\label{fig:img-conv}
	\end{center}
	\vskip -0.2in
\end{figure*}

\clearpage
\twocolumn

\section{Theorems and Proofs}
\subsection{PCA and Linear Autoencoders}\label{sec:appendix-pca}
We show that PCA can be understood as emerging as an edge case from autoencoders, placing it into the same family of dimensionality reduction techniques. The proof up to the consideration of weight decay can in this or a similar form also be found in \citet{pca-ae, pca-ae-2}. We repeat the complete argument here for the reader.

\begin{theorem}
\label{thm:pca-autoencoders}
Let $X$ be a zero-centered dataset whose first $l$ singular values are strictly larger than the $l+1$-st. Further, denote by $\mathcal S$ the set of autoencoders that have linear encoder and decoder without biases, bottleneck dimension $l$ and achieve optimal reconstruction loss on $X$. Then an autoencoder $(E, D) \in \mathcal S$ learns PCA (up to a rotation or reflection) if and only if it also achieves minimal weight decay loss among the autoencoders in $\mathcal S$.
\end{theorem}
\begin{prf}
Let $E \in \mathbb R^{l \times n}$ be the linear encoder, $D \in \mathbb R^{n \times l}$ the linear decoder. Furthermore, let $W \in \mathbb R^{l \times n}$ be the PCA solution, i.e., the matrix whose $l$ rows are the first $l$ principal components. By assumption about $X$'s singular values, the set of the first $l$ principal components and thus the subspace $V'\subset \mathbb{R}^n$ that they span are unique.

Let $X \in \mathbb R^{n, m}$ be the data matrix and set \mbox{$\mathcal W = \{W' \in \mathbb R^{l, n} \; | \; W' \text{ has orthonormal rows}\}$}.
Then the PCA objective (up to rotation and reflection) can be written as 
\begin{equation}
\label{eq:pca-objective}
W \in \argmin_{W' \in \mathcal W} \| X - W'^TW'X \|_2^2.
\end{equation}
The autoencoder objective is given by
\begin{equation}
\label{eq:ae-objective}
E, D = \argmin_{E' \in \mathbb R^{l,n}, D' \in \mathbb R^{n,l}} \| X - D'E'X \|_2^2.
\end{equation}
We want to argue that modulo a multiplication by an invertible matrix in latent space, the two objectives defined in Equations~\eqref{eq:pca-objective} and~\eqref{eq:ae-objective} agree.

\begin{claim}
The possible autoencoder solutions are precisely those matrices of the form $(E, D) = (AW, W^TA^{-1})$ for an $A \in \text{GL}(l, \mathbb R)$.
\end{claim}
\begin{claimproof}
First, we want to show that the PCA solution minimizes the autoencoder objective. If we can show this, one implication follows trivially. Note that the image of $DE$ is an $l$-dimensional subspace $V$ of the vector space $\mathbb R^n$, so that the autoencoder objective in Equation~\eqref{eq:ae-objective} comes down to mapping the dataset into $V \subset \mathbb R^n$ while minimizing the $\ell_2$ distance between a data point and its image. In the Hilbert space $\mathbb R^n$, the minimality condition implies that such a map must be given by an orthogonal projection, independent of the subspace we project onto. In other words, let $B\in \mathbb{R}^{l, n}$ be a matrix whose rows are an orthonormal basis of $V$. Then the orthogonal projection $\mathbb{R}^n \to V$ is given by $v \mapsto B^TBv$. By the minimality criterion, we have $DE = B^TB$. As $B$ is feasible for the PCA objective, the PCA solution also solves the autoencoder objective.

Second, we have to show that every autoencoder solution $(E, D)$ is of the form $(E, D) = (AW, W^TA^{-1})$ for an $A \in \text{GL}(l, \mathbb R)$.
The orthogonal projection to $V'$, the space given by the first $l$ principal components, is given by $W^T W$. By our argument above and the uniqueness of $V'$, the matrix $DE$ must equal $W^TW$. Since $W$ has orthogonal rows, it is surjective and thus $D$ and $W^T$ must have the same row space. In other words, there is some $A\in \text{GL}(l, \mathbb{R})$ such that $D = W^TA^{-1}$. Orthogonality of the rows of $W$ implies $WW^T = I$. Multiplying $DE = W^TW$ by $W$ from the left yields
\begin{align}
    A^{-1}E &= WW^TA^{-1}E = W(DE) = W(W^TW) \\
    &= W.
\end{align}

\end{claimproof}

In the following, we find further restrictions on the matrix $A$ resulting from weight decay on $E$ and $D$.
\begin{claim}
If $E$ and $D$ have minimal Frobenius norm among all $(E, D)\in \{(AW, W^TA^{-1}) | A\in \text{GL}(l, \mathbb{R})\}$, then the additional matrix $A$ is a rotation or reflection.
\end{claim}
\begin{claimproof}
First recall that for a real matrix $M \in \mathbb R^{m, n}$, the Frobenius norm is $\| M \|_F^2 = \tr (MM^t)$.
Consequently the Frobenius-Norm is invariant under multiplication by an orthogonal matrix.
Second, recall that the only freedom $E$ and $D$ have lies in the additional invertible matrix $A$.
Performing an SVD of $A$, we obtain $A = U^t \Sigma V$ where $U$ and $V$ are $l \times l$ orthogonal and \mbox{$\Sigma = (\sigma_1, \sigma_2, ..., \sigma_{l-1}, \sigma_l)$} is $l \times l$ diagonal with $\sigma_i \neq 0$. This allows to evaluate the Frobenius norm of the decoder $D$ as
\begin{equation}
    \begin{split}
        \| D \|_F &= \| W^T A^{-1} \|_F \\
        &= \| W^T V^T \Sigma^{-1} U \|_F \\
        &= \| \Sigma^{-1} \|_F,
    \end{split}
\end{equation}
where we used that $U$, $V$ and $W$ all have orthonormal rows. Analogously, we obtain
\begin{equation}
    \begin{split}
        \| E \|_F = \| \Sigma \|_F.
    \end{split}
\end{equation}
This shows that
\begin{equation}\label{eq:objective}
\begin{split}
\text{loss}_{\text{weight decay}} &= \| D \|_F^2 + \| E \|_F^2 \\
&= \| \Sigma^{-1} \|_F^2 + \| \Sigma \|_F^2 \\
&= \sum_i \sigma_i^2 + \sigma_i^{-2},
\end{split}
\end{equation}
which is minimal if and only if
\begin{equation}
    \sigma_i = \pm 1.
\end{equation}
Consequently, weight decay restricts the autoencoders degree of freedom to a matrix of the form
\begin{equation}
    A = U^t \diag(\pm 1) V \in O(l),
\end{equation}
which is orthogonal since $U$, $\diag(\pm 1)$ and $V$ are so.
\end{claimproof}

This shows that the autoencoder differs from PCA only by a rotation and/or reflection, which completes the proof.
\end{prf}
We believe that Theorem~\ref{thm:pca-autoencoders} closely applies in practice, where autoencoders are typically trained to minimize a weighted sum of reconstruction and weight decay loss, as long as the weight of the regularizer is reasonably small.

\subsection{Pullback Metric in Coordinates}\label{sec:appendix-coordinates}
The pullback of the Euclidean metric under the decoder $D$ takes a very simple form in coordinates:
\begin{theorem}[Pullback Metric in Coordinates]
Given a point $p \in \mathbb R^l$, the pullback metric at $p$ in coordinates is
\begin{equation}
    \langle \cdot, \cdot \rangle_p \coloneqq D^*g_{e_p} = \left( J_pD \right)^t J_pD \in \mathbb R^{2, 2},
\end{equation}
where $J_pD$ is the Jacobian of the decoder at $p$.
\end{theorem}
\begin{prf}
After choosing coordinates on $\mathbb R^l$ and $\mathbb R^n$, the differential of $D$ at $p$ becomes the Jacobian matrix which we denote by $J_pD$, and the inner product in latent space $(\mathbb R^l, D^*g)$ is given by
\begin{align}\label{eq:pullback-metric}
\langle v, w \rangle_p &\coloneqq D^*g_p(v, w) = d_pD(v)^t d_pD(w)\\
&= \left( J_pD v\right)^t  J_pD  w = v^t  \left(J_pD^t  J_pD\right)  w.
\end{align}
\end{prf}

\subsection{Properties of the Determinant Regularization Objective}\label{sec:appendix-invariance}
In this subsection, we prove properties of our secondary objective defined in Equation~\eqref{eq:geomreg-objective}, in particular its minimum and some of its invariances.

\begin{theorem}\null\hfill
\label{thm:prop_reg}
    \begin{enumerate}[leftmargin=15pt, labelindent=0pt, itemsep=1pt, parsep=1pt, topsep=1pt, partopsep=1pt]
        \item $\mathcal{L}_{\det}(D) \geq 0 $
        \item $\mathcal{L}_{\det}(D) = 0$ if and only if  for all $ x, x' \in X$ we have \mbox{$ \det((J_{E(x)}D)^tJ_{E(x)}D) = \det ((J_{E(x')}D)^tJ_{E(x')}D)$.}
        \item If $D$ and $\tilde{D}$ are two decoders and there is some $c>0$ such that for all $x\in X$ we have $\det((J_{E(x)}D)^tJ_{E(x)}D) = c \det((J_{E(x)}\tilde{D})^tJ_{E(x)}\tilde{D})$, then $\mathcal{L}_{\det}(D) = \mathcal{L}_{\det}(\tilde{D})$.
        \item Let $F:\mathbb{R}^l \to \mathbb{R}^l$ be a scaled area--preserving diffeomorphism, so that $\det(J_zF)$ is a constant in $z$. Then the autoencoder $(E, D)$ and the map given by $(F^{-1}\circ E, D\circ F)$ have the same output, the same reconstruction loss, and the same geometric regularizer value $\mathcal{L}_{\det}(D) = \mathcal{L}_{\det}(D\circ F)$.
    \end{enumerate}
\end{theorem}
\begin{prf}\null \hfill
\begin{enumerate}[align=left, leftmargin=0pt, labelindent=\parindent,
listparindent=\parindent, labelwidth=0pt, itemindent=!]
    \item Variances are non-negative.
    \item The ``if'' part is clear since the variance of a constant is zero. The ``only if'' part follows since all data samples have equal probability mass in each batch and batches are also collected uniformly from the whole dataset.
    \item Analogous to  4.
    \item Since the applications of $F$ and $F^{-1}$ cancel, the outputs and reconstruction losses agree.
    
    Let $d=\det(J_zF)$ be the determinant of $F$'s Jacobian. For $x\in X$, we have
    \begin{align}
        J_{F^{-1}(E(x))}(D\circ F) = J_{E(x)}(D) J_{F^{-1}(E(x))}(F)
    \end{align}
    and so
    \begin{align}
        &\det\left(J_{F^{-1}(E(x))}(D\circ F)^tJ_{F^{-1}(E(x))}(D\circ F)\right) \\
        &\quad= d^2 \det\left(J_{E(x)}(D)^tJ_{E(x)}(D)\right).
    \end{align}
    Thus, the value of the regularizer remains unchanged
    \begin{align}
        &\mathcal{L}_{\det}(D\circ F) \\
        &= \Varl_{x\sim\mathcal{U}(B)}\left(\log\left(\det\left( J_{F^{-1}(E(x))}(D\circ F)^t\right.\right.\right.\\
        &\qquad\qquad\qquad\qquad \;\, \left. \left.\left.J_{F^{-1}(E(x))}(D\circ F)\right)\right)\right) \\
        &=\Varl_{x\sim\mathcal{U}(B)}\left(\log\left(d^2 \det(J_{E(x)}(D)^tJ_{E(x)}(D))\right)\right)\\
        &=\Varl_{x\sim\mathcal{U}(B)}\left(\log(\det(J_{E(x)}(D)^tJ_{E(x)}(D)))\right. \\
        &\qquad\qquad\qquad \;+ 2\log(d)\big)\\
        &=\Varl_{x\sim\mathcal{U}(B)}\left(\log(\det(J_{E(x)}(D)^tJ_{E(x)}(D)))\right)\\
        &=\mathcal{L}_{\det}(D).
    \end{align}
\end{enumerate}
\end{prf}

The first two parts of Theorem~\ref{thm:prop_reg} show that our regularizer becomes minimal exactly when the decoder is area-preserving at the embedding points. The second two parts describe invariances of our regularizer. Our regularizer is insensitive to area-preserving changes of the decoder, or equivalently, of the embedding. This implies scale invariance, a useful property for visualization: 

\begin{corollary}[Scale Invariance]
Let the first layer of the decoder scale by a factor of $\beta \in \mathbb R \setminus \{0\}$, and the embedding by $\beta^{-1}$. Not only does this fix the primary objective (the reconstruction loss), but also our secondary geometric objective.
\end{corollary}
\begin{prf}
This is a special case of Theorem~\ref{thm:prop_reg} in which $F$ is the multiplication with $\beta$.
\end{prf}
In particular, our regularizer does not favor decoders with Jacobian of small norm, in contrast to the regularizer of~\citet{chen2020learning}, see \cite{lee2022regularized}.

\section[Relation to Lee et al., 2022]{Relation to \citet{lee2022regularized}}
\label{sec:appendix-relation-lee}

\citet{lee2022regularized} describe a hierarchy of geometry-preserving mappings consisting, from strongest geometry-preservation to weakest, of isometries, scaled isometries, conformal maps and area-preserving maps. Their proposed regularizers tackle the case of scaled isometries and they explicitly refrain from exploring area-preserving maps.

In turn, our regularizer promotes area-preservation. We will first explain how the functional form of our regularizer differs from that of~\citet{lee2022regularized} and then discuss how our regularizer achieves a similar goal as that of \citet{lee2022regularized} in practice.

Denote the determinant of the pullback metric at point $z$ by $d(z) \coloneqq \det((J_zD)^tJ_zD)$ and by $\lambda_1(z), \dots, \lambda_l(z)$ the eigenvalues of the pullback metric at $z$. In slight abbuse of notation, we will write $z\sim \mathcal{U}(B)$ when we mean $z=E(x)$ and $x$ being sampled uniformly from the batch $x\sim\mathcal{U}(B)$.

Rewriting our regularizer from Equation~\eqref{eq:geomreg-objective}, we get
\begin{align}
    &\mathcal L_{\det} \\
    &= \Varl_{z \sim \mathcal U(B)} \left[ \log \left(d(z)\right)\right] \\
    & = \El_{z \sim \mathcal U(B)}\left[ \left(\log(d(z))- \El_{z' \sim \mathcal U(B)}\log(d(z'))\right)^2\right]\\
    &= \El_{z \sim \mathcal U(B)}\left[
        \left(\sum_{i=1}^l \log(\lambda_i(z)) \right.\right.\\
        &\qquad\qquad\qquad- \El_{z'\sim\mathcal{U}(B)} \sum_{j=1}^l \log(\lambda_j(z'))\Bigg)^2 \Bigg]\label{eq:geomreg_alternative}
\end{align}
\citet{lee2022regularized}'s regularizer requires the choice of a probability distribution $\mathbb{P}$ on latent space, a map $h: \mathbb{R} \to [0, \infty)$ with $h(1) = 0$ and $h'(\lambda)=0$ if and only if $\lambda=1$ and finally a symmetric map $S:\mathbb{R}^l \to \mathbb{R}$ with 
\begin{align}
    S(\alpha \lambda_1, \dots, \alpha\lambda_l) &= \alpha S(\lambda_1, \dots, \lambda_l)\\
    S(1, \dots, 1) &= 1.
\end{align}
Given $\mathbb{P}$, $h$ and $S$, the regularizer is
\begin{align}
    &\mathcal{L}_{\text{Lee}}(\mathbb{P}, h , S) =\\
    &\quad\El_{z\sim \mathbb{P}}\left(\sum_{i=1}^l h\left(\frac{\lambda_i(z)}{\mathbb{E}_{z'\sim \mathbb{P}}  S(\lambda_1(z'), \dots, \lambda_l(z'))} \right)\right)
\end{align}
The following admissible choices yield the closest match to our regularizer:
\begin{align}
    \mathbb{P}\text{ to be given by }z&=E(x), x\sim \mathcal{U}(B)\\
    h(\lambda) &= \log(\lambda)^2 \\
    S(\lambda_1, \dots, \lambda_l) &= \frac{1}{l}\sum_{i=1}^l \lambda_i 
\end{align}
While they are not the main choices employed by~\citet{lee2022regularized}), they yield the regularizer
\begin{align}
    &\mathcal{L}_{\text{Lee}} = \mathbb{E}_{z \sim \mathcal U(B)}\Bigg[ \sum_{i=1}^l \Bigg(\log(\lambda_i(z)) \\
    &\qquad\qquad\qquad - \log\Bigg(\frac{1}{l}\sum_{j=1}^l \mathbb{E}_{z' \sim \mathcal U(B)} \lambda_i(z')\Bigg)\Bigg)^2\Bigg]\label{eq:lee-reg-alternative}
\end{align}
Comparing Equations~\eqref{eq:geomreg_alternative} and~\eqref{eq:lee-reg-alternative}, the difference between our regularizer and \citet{lee2022regularized}'s amounts to a different ordering of taking expectation, logarithm, sum and square. 

Practically, this means that our method promotes only scaled area-preservation instead of a scaled isometry, see Theorem~\ref{thm:prop_reg}. While we only regularize towards scaled area-preservation, we observe qualitatively in Figure~\ref{fig:indicatrices} that it also seems to favor isotropic decoders. \citet{lee2022regularized} propose the pullback metric determinant's $2$-norm condition number as a measure for the decoder's isotropy. It is equivalent to the ratio of the length of the indicatrices' main axes; an isotropic decoder would have a condition number of one. We calculate the condition number on a regular, sufficiently dense grid in latent space intersected with the datasets convex hull for MNIST. The result is shown in Table~\ref{table:isotropy}. Indeed, the geometric autoencoder is the most isotropic, and the vanilla autoencoder the least (note the huge standard deviation).

\begin{table}[t]
\caption{The pullback metric's $2$-norm condition number acts as a measure for the decoder's isotropy. We evaluate it on the intersection between a regular grid of $100 \times 100$ points and the embedding's convex hull. The results are for the MNIST dataset and we report average and standard deviation over five different initializations. The last two rows refer to the convolutional autoencoders described in Section~\ref{sec:appendix-conv}.}
\label{table:isotropy}
\vskip 0.15in
\begin{center}
\begin{small}
\begin{sc}
\begin{tabular}{lc}
\toprule
Model & Mean Condition Number \\
\midrule
Geom AE (ours) & $2.5 \pm 1.1$ \\
Vanilla AE & $100.0 \pm 900.0$ \\
Topo AE    & $10.1 \pm 22.1$ \\
UMAP AE    & $3.6 \pm 3.1$ \\
PCA        & $1.0 \pm 0.0$ \\
\midrule
Conv Vanilla AE & $14.6 \pm 13.8$ \\
Conv GeomAE (ours) & $2.5 \pm 1.2$ \\
\bottomrule
\end{tabular}
\end{sc}
\end{small}
\end{center}
\vskip -0.1in
\end{table}

\section{Datasets, Metrics, Training}
\subsection{Datasets}
\label{sec:appendix-datasets}
We evaluate the models using the image datasets MNIST~\cite{mnist} and FashionMNIST~\cite{fashion-mnist}, both of which we normalize to the unit interval as proposed by~\citet{topoae}, as well as the single-cell datasets Zilionis~\cite{zilionis}, CElegans~\cite{celegans} and PBMC~\cite{pbmc}.

\begin{figure*}[ht]
\vskip 0.2in
\begin{center}
    \centering
     \begin{subfigure}[b]{0.15\textwidth}
         \centering
         \includegraphics[width=\linewidth]{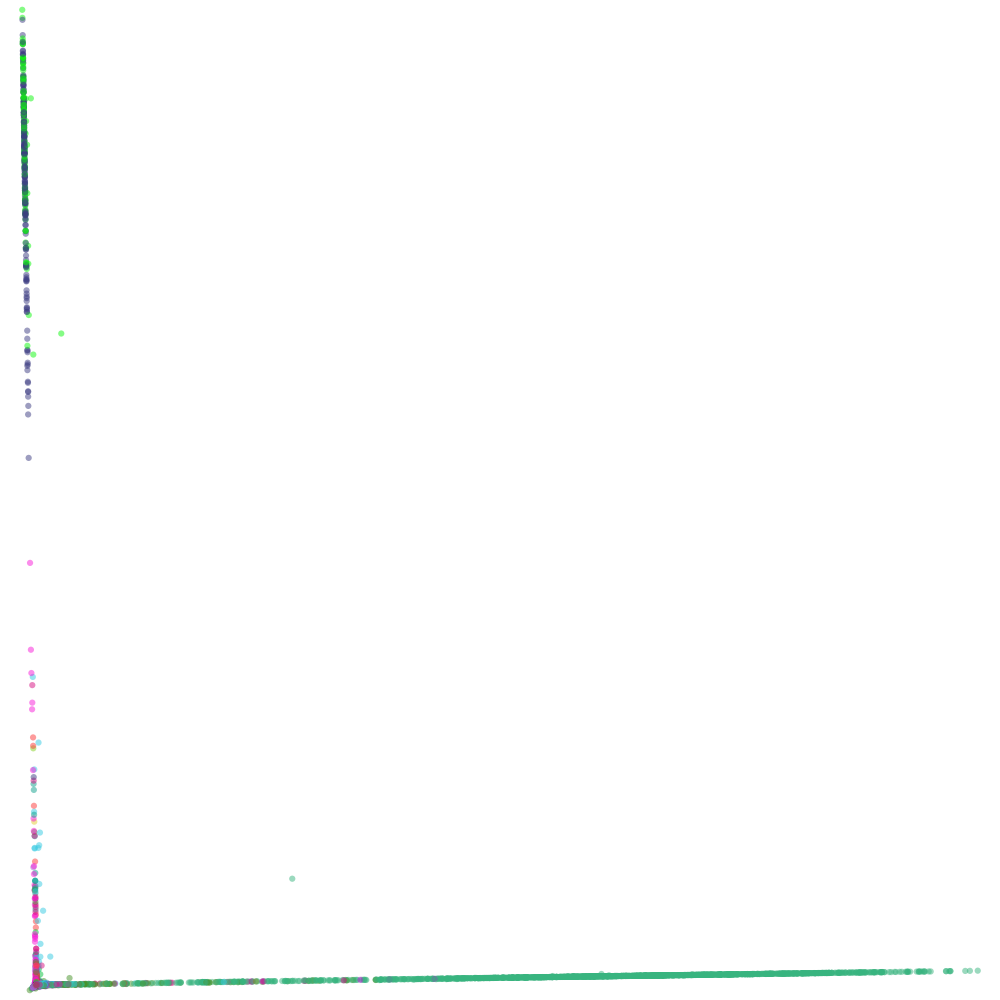}
         \caption{}
         \label{fig:zilionis-raw}
     \end{subfigure}
     \begin{subfigure}[b]{0.15\textwidth}
         \centering
         \includegraphics[width=\linewidth]{media/zilionis/pca/latents.png}
         \caption{}
         \label{fig:zilionis-normalized}
     \end{subfigure}
     \hspace{5mm}
     \begin{subfigure}[b]{0.15\textwidth}
         \centering
         \includegraphics[width=\linewidth]{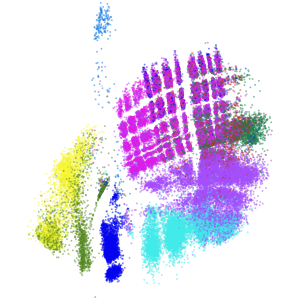}
         \caption{}
         \label{fig:pbmc-artefacts}
     \end{subfigure} 
     \begin{subfigure}[b]{0.15\textwidth}
         \centering
         \includegraphics[width=\linewidth]{media/pbmc/geomreg/latents.png}
         \caption{}
         \label{fig:pbmc-no-artefacts}
     \end{subfigure}  

    \vspace{5mm}
    \begin{subfigure}[b]{\textwidth}
     \begin{subfigure}[b]{0.12\textwidth}
        \centering
             \begin{tikzpicture}[font=\tiny]
                \node[minimum width=\textwidth] {Geom AE};
             \end{tikzpicture}
     \end{subfigure}
     \hspace{.5mm}
     \begin{subfigure}[b]{0.12\textwidth}
        \centering
             \begin{tikzpicture}[font=\tiny]
                \node[minimum width=\textwidth] {Vanilla AE};
             \end{tikzpicture}
     \end{subfigure}
     \hspace{.5mm}
     \begin{subfigure}[b]{0.12\textwidth}
        \centering
             \begin{tikzpicture}[font=\tiny]
                \node[minimum width=\textwidth] {Topo AE};
             \end{tikzpicture}
     \end{subfigure}
     \hspace{.5mm}
     \begin{subfigure}[b]{0.12\textwidth}
        \centering
         \begin{tikzpicture}[font=\tiny]
            \node[minimum width=\textwidth] {UMAP AE};
         \end{tikzpicture}
     \end{subfigure}
     \hspace{.5mm}
     \begin{subfigure}[b]{0.12\textwidth}
        \centering
             \begin{tikzpicture}[font=\tiny]
                \node[minimum width=\textwidth] {PCA};
             \end{tikzpicture}
     \end{subfigure}
     \hspace{.5mm}
     \begin{subfigure}[b]{0.12\textwidth}
        \centering
         \begin{tikzpicture}[font=\tiny]
            \node[minimum width=\textwidth] {$t$-SNE};
         \end{tikzpicture}
     \end{subfigure}
     \hspace{.5mm}
     \begin{subfigure}[b]{0.12\textwidth}
        \centering
         \begin{tikzpicture}[font=\tiny]
            \node[minimum width=\textwidth] {UMAP};
         \end{tikzpicture}
     \end{subfigure}

    \begin{subfigure}[b]{0.12\textwidth}
         \centering
         \includegraphics[width=\linewidth]{media/pbmc/artefacts/geomreg/latents.png}
     \end{subfigure}
     \hspace{.5mm}
     \begin{subfigure}[b]{0.12\textwidth}
         \centering
         \includegraphics[width=\linewidth]{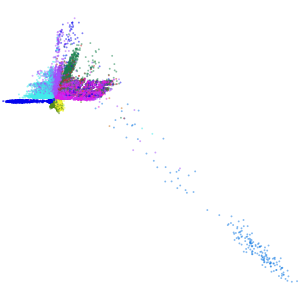}
     \end{subfigure}
     \hspace{.5mm}
     \begin{subfigure}[b]{0.12\textwidth}
         \centering
         \includegraphics[width=\linewidth]{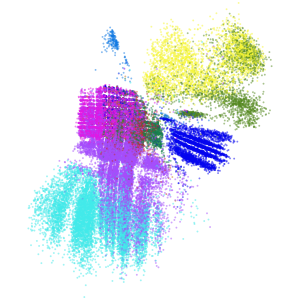}
     \end{subfigure}
     \hspace{.5mm}
     \begin{subfigure}[b]{0.12\textwidth}
         \centering
         \includegraphics[width=\linewidth]{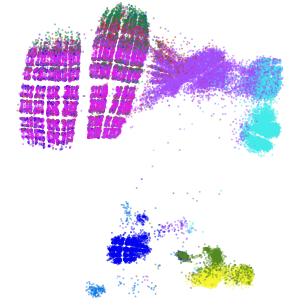}
     \end{subfigure}
     \hspace{.5mm}
     \begin{subfigure}[b]{0.12\textwidth}
         \centering
         \includegraphics[width=\linewidth]{media/pbmc/pca/latents.png}
     \end{subfigure}
     \hspace{.5mm}
     \begin{subfigure}[b]{0.12\textwidth}
         \centering
         \includegraphics[width=\linewidth]{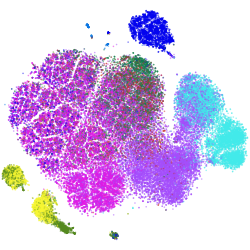}
     \end{subfigure}
     \hspace{.5mm}
     \begin{subfigure}[b]{0.12\textwidth}
         \centering
         \includegraphics[width=\linewidth]{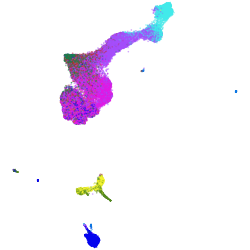}
     \end{subfigure}
     \caption{}
    \label{fig:pbmc-disguised}
    \end{subfigure}
    
    \caption{\textbf{First row:} Our changes in the preprocessing of the Zilionis and the PBMC dataset. Panel~(\subref{fig:zilionis-raw}) shows the PCA embedding of the original preprocessed Zilionis data, which is dominated by two features. Normalizing each feature yields the PCA in Panel~(\subref{fig:zilionis-normalized}). Panel~(\subref{fig:pbmc-artefacts}) shows the geometric autoencoder's embedding of the original preprocessed PBMC dataset, with artefacts originating from the preprocessing. Redoing the preprocessing ourselves yields the Geometric autoencoder embedding in Panel~(\subref{fig:pbmc-no-artefacts}).  \textbf{Second row:} In Panel~(\subref{fig:pbmc-disguised}) we show how different models spot the artefacts in preprocessed PBMC differently well. While the geometric autoencoder exposes the regular structure, UMAP, Topo AE, PCA, and the vanilla autoencoder disguise it almost completely.}
\label{fig:weird-datasets}
\end{center}
\vskip -0.2in
\end{figure*}

The single-cell datasets where obtained from \url{http://cb.csail.mit.edu/cb/densvis/datasets/}, where they are already preprocessed~\cite{data-url}. On Zilionis, we additionally normalize each feature to have a mean of zero and a standard deviation of one. This is necessary since two features dominated the dataset, as indicated by the PCA embedding of the non-normalized dataset (see Figure~\ref{fig:zilionis-raw},~\subref{fig:zilionis-normalized}). Our analysis furthermore revealed an artefact in the preprocessed PBMC data, see Figure~\ref{fig:pbmc-artefacts}. We downloaded the original data from \url{https://support.10xgenomics.com/single-cell-gene-expression/datasets} and preprocessed the dataset following the procedure in~\citet{Kobak}: First, we selected the 1000 most variable genes. Then, we normalized the library sizes to the median library size in the dataset. Next, we log-transformed with $\text{log}_2 (x + 1)$, and finally applied PCA to reduce that dataset to $50$ dimensions. This second version of the PBMC dataset did not show the artefact anymore, see Figure~\ref{fig:pbmc-no-artefacts}.

We created the Earth dataset using the python package ``mpl\_toolkits''. It consists of $100\,000$ points randomly sampled from the $S^2$, wherever there is landmass on earth, excluding Antarctica. Every point is labeled by the continent to which it belongs. Furthermore, in the python package we used, Europe and Russia have the same label. The labelled datasets can be found in Figure~\ref{fig:labels}.

\subsection{Training}
\label{sec:appendix-training}
All of the autoencoders except for the UMAP autoencoder are optimized using ADAM~\cite{adam}, and trained using a batch size of $125$, learning rate $10^{-3}$ and a weight decay of $10^{-5}$. For the UMAP autoencoder, we used the standard settings with an additional weight decay of $10^{-5}$. The vanilla, topological and geometric autoencoders are trained for $100$ epochs.
For the UMAP autoencoder we use the standard settings of the TensorFlow implementation, which in particular trains for only one epoch. However, an epoch of UMAP autoencoder is much longer than an epoch of the other autoencoders. Usually, an epoch iterates once over the entire dataset. The UMAP autoencoder iterates over pairs of datapoints that are incident in the $k$NN graph. Additionally, this graph is upsampled to reflect a weighing of the $k$NN graph. We measured the size of the resulting dataset. For MNIST experiment, we found it to be more than $2600$ times larger than the normal MNIST dataset. Therefore, a single epoch of UMAP autoencoder corresponds to even more than $100$ epochs, which we use for the other autoencoder models. An overview for all datasets can be found in Table~\ref{table:pumap-epochs}.


\begin{table}[t]
\caption{In the standard implementation of Parametric UMAP, the number of training samples (\# samples) varies from the datasets' size (\# dataset). This table provides an overview of how many ``actual epochs`` one ``Parametric UMAP epoch`` corresponds to.}
\label{table:pumap-epochs}
\vskip 0.15in
\begin{center}
\begin{small}
\begin{sc}
\begin{tabular}{lllc}
\toprule
Dataset & \# Dataset & \# Samples & Epochs \\
\midrule
MNIST        & $70\,000$ & $182\,278\,632$  & $2600$ \\
FashionMNIST & $70\,000$ & $191\,104\,176$  & $2800$ \\
PBMC         & $68\,551$ & $191\,021\,188$  & $2800$ \\
CElegans     & $86\,024$ & $236\,600\,132$ & $2800$ \\
Zilionis     & $48\,969$ & $142807948$  & $3000$ \\
\bottomrule
\end{tabular}
\end{sc}
\end{small}
\end{center}
\vskip -0.1in
\end{table}

\subsection{Implementation}

The vanilla and the geometric autoencoder were implemented by us in PyTorch. In data loading, training schedule and quantitative evaluation we follow the PyTorch implementation of the topological autoencoder\footnotemark, as referenced in ~\citet{topoae}. For the differential geometry involved, we use the Geomstats package~\cite{geomstats}.

We plot the indicatrices on a regular grid in latent space, intersected with the convex hull of the dataset. We furthermore scale them globally such that the elongated ones do not cover each other too heavily. This is justified because we only care about the relative size of the indicatrices, not their absolute size.

\footnotetext{https://github.com/BorgwardtLab/topological-autoencoders}

Table~\ref{table:runtimes} summarizes the training runtimes on MNIST, averaged over five random initializations.

\begin{figure}[ht]
\vskip 0.2in
\begin{center}
    \centering
     \includegraphics[width=\linewidth]{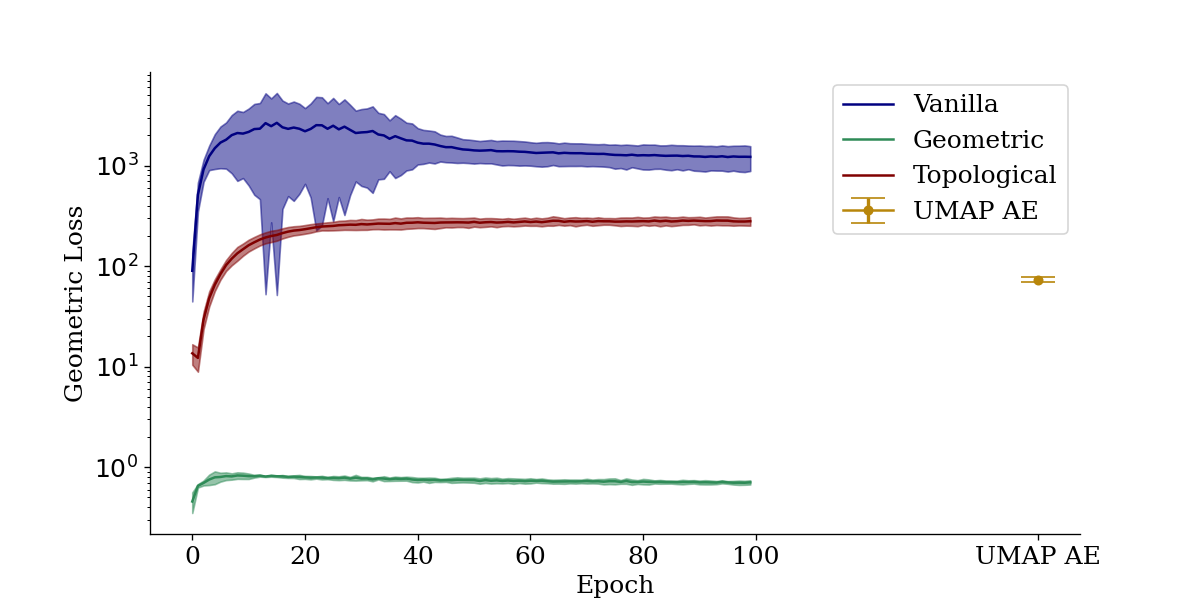}
    \caption{The geometric loss curves for the autoencoder models.}
\label{fig:geometric-loss}
\end{center}
\vskip -0.2in
\end{figure}

\begin{table}[t]
\caption{Training runtimes on MNIST, averaged over five random initializations. Including one standard-deviation.}
\label{table:runtimes}
\vskip 0.15in
\begin{center}
\begin{small}
\begin{sc}
\begin{tabular}{lc}
\toprule
Model & Time [min] \\
\midrule
Geom AE    & $39.00 \pm 1.00$ \\
Vanilla AE & $24.40 \pm 0.40$ \\
Topo AE    & $82.10 \pm 2.10$ \\
UMAP AE\footnotemark    & $108.40 \pm 4.9$ \\
PCA        & $00.20 \pm 0.01$ \\
$t$-SNE    & $05.70 \pm 0.50$ \\
UMAP       & $03.51 \pm 0.03$ \\
\bottomrule
\end{tabular}
\end{sc}
\end{small}
\end{center}
\vskip -0.1in
\end{table}

\footnotetext{Even though we train with the default batch size of $1000$, the speed test uses $125$ for consistency.}
Figure~\ref{fig:geometric-loss} shows the geometric loss curve for the vanilla autoencoder, the topological autoencoder and the geometric autoencoder, averaged over $4$ random initialization. The geometric loss of the geometric autoencoder is roughly three orders of magnitude smaller than that of the vanilla and the topological autoencoder. It furthermore decreases with training, as opposed to the geometric loss of the topological autoencoder. We do not show the graph for the UMAP autoencoder, since it comes pre-implemented in TensorFlow, and our geometric loss is implemented in PyTorch. To compute diagnostics of the final model, we transferred the trained network weights from TensorFlow to PyTorch.

We use the automatic differentiation abilities of PyTorch, namely the function \textit{jacrev} in the \textit{functorch} library, for efficiently calculating the decoders pullback metric (see Algorithm~\ref{alg:gen_jac_det}). Since we are using ELU activations, the decoder $D \colon \mathbb R^2 \to \mathbb R^n$ is differentiable everywhere, and thus has a well defined Jacobian matrix $J_xD \in \mathbb R^{n \times 2}$ for all $x \in \mathbb R^2$.


\begin{algorithm}[tb]
   \caption{Calculating the Generalized Jacobian Determinant}
    \label{alg:gen_jac_det}
    \begin{algorithmic}
    \STATE from \textbf{functorch} import \textbf{jacrev}
    \STATE import \textbf{torch}
    \STATE
    \STATE $J = $ \textbf{jacrev}(decoder)(batch\_of\_embeddings)
    \STATE metric = J.T @ J
    \STATE gen\_jac\_det = torch.det(metric)
    \STATE \textbf{return} gen\_jac\_det
    \end{algorithmic}
\end{algorithm}

\subsection{Metrics}
\label{sec:appendix-metrics}
We evaluate each model on five random seeds. As mentioned in the main paper, we evaluate the embeddings with metrics from~\citet{topoae} and ~\citet{tsne-single-cell}. Namely, there are the three local metrics $\mathit{KL_{0.1}}$, \textit{kNN}, \textit{Trust} and the three global metrics \textit{Stress}, $\mathit{KL_{100}}$, \textit{Spear}.
\begin{itemize}[leftmargin=15pt, labelindent=0pt, itemsep=1pt, parsep=1pt, topsep=1pt, partopsep=1pt]
\item The \textit{kNN} metric measures which ratio of nearest neighbors in the the embedding are also nearest neighbors in the original dataset~\cite{Kobak, parametric-umap}.
\item The \textit{Trust} metric is a metric based on nearest neighbor ranks~\cite{trust}.
\item The \textit{Stress} metric coincides with the loss of multidimensional scaling, and measures the sum of the squared differences of the distances between all pairs of embedding points and the corresponding differences of all pairs of input points~\cite{topoae}.
\item The \textit{Spear} metric measures the Spearman correlation between the distances between all pairs of embedding and input points~\cite{tsne-single-cell}.
\item The $\mathit{KL_{\sigma}}$ metrics ($\sigma = 0.1, 100$) measure the Kullback-Leibler divergence between a density estimate $f_\sigma^X$ of the dataset $X$ and the corresponding estimate $f_\sigma^Z$ of the embedding $Z$. As a density estimate, we use the \textit{distance to a measure} density estimator~\cite{distance-to-a-measure} defined as $f_\sigma^X(x) = \sum_{y \in X} \exp \left( - \sigma^{-1} \frac{\|x-y\|_2^2}{\max_{y', x' \in X} \|y'-x'\|_2^2} \right)$, where the parameter $\sigma$ defines a length scale. Ideally, the $\mathit{KL_{\sigma}}$ value is small, which means that the density estimation in latent space is similar to the density estimation of the actual dataset.
\end{itemize}

The metrics depending on the number of nearest neighbors are averaged over a range of values from $10$ to $200$ in steps of $10$, as proposed by~\citet{topoae}.

\subsection{The Geometric Loss}
\label{sec:appendix-gradients}
In this subsection we further investigate how the geometric loss $\mathcal L_{\det}$ affects encoder and decoder.

In our implementation both encoder and decoder accumulate gradients during one step of backpropagation. This may seem a bit counterintuitive, since at first glance the secondary objective only depends on the decoders' Jacobian matrix, hence only on the decoder (see Equations~\eqref{eq:pullback-metric-coordinates} and~\eqref{eq:geomreg-objective}). The reason for the encoder to accumulate gradients is that the the pullback metric tensor is also function of latent space point at which it is evaluated. Evaluating it at a given embedding, as we do in our geometric objective (Equation~\eqref{eq:geomreg-objective}), gives us the set of pullback metric tensors we use.

Consequently, the autoencoder can reduce the geometric loss in two ways. First, it can change the decoder's weights in order to achieve more uniform contraction for a fixed set of embedding points. Second, it can push around the embedding into areas where the decoder contracts more homogeneously. In practice, a geometric autoencoder will pursue both approaches simultaneously.

A simple gradient stop layer could prevent the encoder from receiving gradients from the geometric regularizer. We consider this an interesting avenue for future work, but believe that such a gradient stop layer might impede the autoencoder's ability to achieve homogeneous stretching of the embedding by the decoder. 

\section{Additional Experiments and Insights}
\subsection{Autoencoders as Orthogonal Projectors}
\label{sec:appendix-orthogonal-projection}
As described in Appendix~\ref{sec:appendix-pca}, a linear autoencoder reduces to PCA. Viewed through the geometric lens, the linear decoder leads to a linear subspace as reconstruction manifold $M$, while the encoder's job is to place the reconstruction of an input point onto this reconstruction manifold. When trained with the usual mean squared error (MSE), the optimal encoder for a given linear decoder projects each input point orthogonally to the linear reconstruction manifold. To see this, consider the sphere around an input point $x$ through some point $z$ on $M$. If the vector $z-x$ is not orthogonal to the linear subspace $M$, the sphere intersects $M$ and there exists some $y \in M$ inside the sphere. So $z$ does not yield optimal mean squared error.
Now, consider the general, non-linear case. If for a given decoder and thus a given reconstruction manifold (not necessarily a linear subspace anymore), there exists a data point $x$ and a point $y \in M$ such that any point $z \in M$ is at least as far from $x$ as $y$, i.e., $\|x-y\| \leq  \|x-z\|$ $\forall z \in M$, then the vector $x-y$ is orthogonal to any tangent vector at $y$ by the same argument as in the PCA case.

Note, however, that there are some subtleties in the non-linear case. For instance, the encoder is limited by its architecture. So it might not be able to express the function that maps each input point in such a way to latent space that their corresponding positions on the reconstruction manifold are the desired orthogonal projections. Moreover, it might not be possible to orthogonally project an input point to the reconstruction manifold in the first place, e.g., if $x = (2,0,...,0)$ and $\text{im}(D)$ is the open unit ball in the first $l$ dimensions of $\mathbb R^n$. Also, there might be multiple closest points to a data point, e.g., if the data point is the center of a sphere and $M$ the sphere.

\subsection{Flexibility of our Regularizer}
\label{sec:appendix-conv}
We expect our regularizer to work for different kinds of autoencoder models, as long as the decoder is differentiable (at least almost everywhere). As an example, we present its effect on convolutional autoencoders.

\textbf{Application to Convolutional Autoencoders}
The proposed regularizer as well as the diagnostic methods also work for convolutional autoencoders. Adapting the architecture proposed by~\citet{topoae} to ensure a two-dimensional latent space, we train convolutional autoencoders on the image datasets MNIST and FashionMNIST. Indeed, the geometric regularizer still ensures more homogeneous contraction (see Figure~\ref{fig:img-conv}). The corresponding quantitative comparisons can be found in Tables~\ref{table:isotropy} and~\ref{table:metrics-convnets}.

\textbf{Pulling Back other Metrics}
In the main paper, we have discussed how pulling back the Euclidean metric from output space yields a metric on latent space which respects the geometry of the reconstruction manifold $M$.

The method we describe, however, is not limited to the Euclidean metric. Rather, it is applicable if the output space is equipped with a Riemannian metric $g'$ or a scalar product. Let such a Riemannian metric $g'$ be represented by a matrix $A \in \mathbb R^{n,n}$. Then it follows directly from the proof of Proposition~\ref{prop:pullback-metric-coordinates}, that the pullback $D^*g'$ is in coordinates given by
\begin{equation}
    D^*{g'}_p = \left(J_pD\right)^t A_D(p) J_pD
\end{equation}
for any point $p \in \mathbb R^l$. In the case of a general scalar product, we can simply pullback the scalar product.

Note that for instance the $\ell^1$ distance on euclidean space is not induced by a scalar product and hence not a Riemannian metric. Nevertheless, it would be possible, albeit somewhat inconsistent, to use the $\ell^1$ metric in the reconstruction loss, but pull back the standard scalar product in Euclidean space. 

\textbf{Training with other Loss Functions}
Even though all the autoencoders considered in this paper are trained with main $\ell^2$ loss, our method is non-restrictive in the choice of loss function. This is because the regularizer depends only on the decoder's architecture. In particular, geometric autoencoders can also be trained with $\ell^1$ or cross-entropy loss.


\subsection{Effect on Downstream Applications}

Since autoencoders are often part of a larger pipeline, a possible question is how the geometric regularizer affects downstream applications. While the focus of our work is on visualization and representations for downstream tasks typically have more than two dimensions, we did compare the class separation in the two-dimensional vanilla and geometric embeddings. First, we cluster both the vanilla and the geometric embedding with HDBSCAN as in~\citet{paper-tuebingen}. We then evaluate these clusters against the existing class labels by computing the adjusted Rand index. Looking at Figure~\ref{fig:embeddings}, we would expect our geometric autoencoder to outperform the vanilla autoencoder, as it creates visually better separated clusters. For HDBSCAN we use the sklearn implementation with default parameters. Table~\ref{table:hdb-scan} shows that our method has a higher score on all datasets and thus outperforms the vanilla model.

\begin{table}[t]
\caption{Adjusted Rand index of class labels after performing an HDBSCAN clustering on the vanilla and the geometric embeddings with default parameters. Averaged over five random initializations of the network. Bold indicates first place.}
\label{table:hdb-scan}
\vskip 0.15in
\begin{center}
\begin{small}
\begin{sc}
\begin{tabular}{lll}
\toprule
Dataset &  Vanilla Score & Geometric Score \\
\midrule
MNIST        & $0.21 \pm 0.09$ & $\mathbf{0.23 \pm 0.15}$ \\
FashionMNIST & $0.18 \pm 0.06$  & $\mathbf{0.21 \pm 0.07}$ \\
PBMC         & $0.047 \pm 0.012$ & $\mathbf{0.074 \pm 0.011}$ \\
CElegans     & $0.036 \pm 0.004$ & $\mathbf{0.095 \pm 0.010}$ \\
Zilionis     & $0.47 \pm 0.13$  & $\mathbf{0.57 \pm 0.14}$ \\
\bottomrule
\end{tabular}
\end{sc}
\end{small}
\end{center}
\vskip -0.1in
\end{table}

\begin{figure*}[ht]
	\vskip 0.2in
	\begin{center}
    \begin{subfigure}{\textwidth}
    \centering
        \begin{subfigure}[b]{0.24\textwidth}
			\centering
			\includegraphics[width=\linewidth]{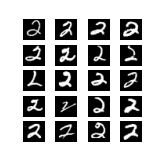}
			\caption{Lower Cluster}
		\end{subfigure}
        \begin{subfigure}[b]{0.24\textwidth}
			\centering
			\includegraphics[width=\linewidth]{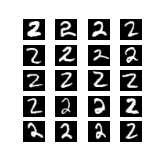}
			\caption{Upper Cluster}
		\end{subfigure}
        \hfill
        \begin{subfigure}[b]{0.24\textwidth}
			\centering
			\includegraphics[width=\linewidth]{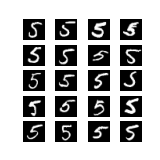}
			\caption{Left Cluster}
		\end{subfigure}
        \begin{subfigure}[b]{0.24\textwidth}
			\centering
			\includegraphics[width=\linewidth]{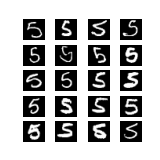}
			\caption{Right Cluster}
		\end{subfigure}
  
    \end{subfigure}
		\caption{Random ground-truth samples from the two clusters of each of the digit $2$ and $5$ in the geometric MNIST embedding (Figure~\ref{fig:embeddings-mnist-geomreg}). While the upper right cluster of the digit $2$ contains mainly straight digits, the lower left cluster contains mostly ones with a loop. For the digit $5$, we observe that the digits corresponding to the left cluster are slanted, while the ones corresponding to the right cluster are not.}
		\label{fig:digit-comparison}
	\end{center}
	\vskip -0.2in
\end{figure*}

\subsection{Semantic Information in MNIST embeddings}
\label{section:appendix-semantic-info}
Comparing the vanilla and the geometric embeddings of MNIST (Figures~\ref{fig:embeddings-mnist-geomreg},\subref{fig:main-evaluation-b}), one can see that the geometric autoencoder creates two clusters for the digit $2$ (pink). We manually inspected these clusters and found that the lower left cluster of the geometric embedding contains digits with loop and curved lower stroke, the upper right cluster contains digits without loop and curved lower stroke (samples can be found in Figure~\ref{fig:digit-comparison}). The vanilla autoencoder is also able to pick up on this signal, but not as well as the geometric autoencoder, because one of the two clusters is in the densely packed region of the vanilla autoencoder embedding and thus barely visible. Hence, the geometric autoencoder is able to pick up and display some semantic better that the vanilla autoencoder. In Figure~\ref{fig:main-evaluation-c}, one can see that the topological autoencoder also creates two clusters for the digit $2$. Investigating them more closely we find that they too differentiate between more curly and more straight $2$'s. Similar to the vanilla autoencoder, the second cluster for the topological autoencoder is in the dense region of the embedding, so that it is hard to spot.
\\
A similar phenomenon holds for the digit $5$. The geometric autoencoder differentiates between slanted digits in the left cluster and straight digits in the right cluster. Again, so does the vanilla autoencoder, but due to the heavy overlapping in the contracted area we can only see one cluster.

\onecolumn
\begin{sidewaystable}[t]
\caption{Table underlying the aggregated metrics of Table~\ref{table:aggregated-metrics}. Additionally contains the reconstruction loss wherever existent (MSE). Averaged over five runs, bold+underlined indicates first, bold second place. The arrows point to the desirable direction of each metric. ConvNet results are considered separately, see Appendix~\ref{sec:appendix-conv} and Table~\ref{table:metrics-convnets}.}
\label{table:metrics}
\vskip 0.15in
\begin{center}
\begin{small}
{\tiny
\begin{sc}
\begin{tabularx}{\textwidth}{ll|lll|lllll}
\toprule
 \multicolumn{2}{c|}{} & \multicolumn{3}{c|}{Local} & \multicolumn{4}{c}{Global} & \\
 \midrule
Dataset & Model & $\dkl_{0.1}$ ($\downarrow$) & kNN ($\uparrow$) & Trust ($\uparrow$) & Stress ($\downarrow$) & $\dkl_{100}$ ($\downarrow$) & Spear ($\uparrow$) & MSE ($\downarrow$) \\
\midrule

{} & Geom AE (ours)         &                        0.169 $\pm$ 0.023 &                          0.356 $\pm$ 0.007 &                          0.938 $\pm$ 0.003 &                 \textbf{6.2 $\pm$ 1.2} &                        2.2e-07 $\pm$ 3e-08 &                         0.4 $\pm$ 0.02 &                        0.0357 $\pm$ 0.0003 \\
{} & Vanilla AE      &               \textbf{0.133 $\pm$ 0.007} &                           0.322 $\pm$ 0.01 &                           0.93 $\pm$ 0.004 &                             11 $\pm$ 3 &                        1.8e-07 $\pm$ 2e-08 &                        0.44 $\pm$ 0.05 &               \textbf{0.0356 $\pm$ 0.0007} \\
{} & Topo AE        &   \underline{\textbf{0.094 $\pm$ 0.003}} &                          0.311 $\pm$ 0.005 &                          0.925 $\pm$ 0.002 &                        8.91 $\pm$ 0.05 &   \underline{\textbf{9.3e-08 $\pm$ 6e-09}} &   \underline{\textbf{0.64 $\pm$ 0.01}} &                        0.03701 $\pm$ 8e-05 \\
MNIST & UMAP AE &                         0.18 $\pm$ 0.007 &   \underline{\textbf{0.4104 $\pm$ 0.0011}} &   \underline{\textbf{0.9483 $\pm$ 0.0003}} &                          7.3 $\pm$ 0.6 &                        3.1e-07 $\pm$ 2e-08 &                        0.34 $\pm$ 0.02 &   \underline{\textbf{0.0335 $\pm$ 0.0003}} \\
{} & UMAP           &                         0.19 $\pm$ 0.002 &                        0.4013 $\pm$ 0.0003 &             \textbf{0.94638 $\pm$ 0.00051} &   \underline{\textbf{4.79 $\pm$ 0.03}} &                        4.1e-07 $\pm$ 1e-08 &                    0.3377 $\pm$ 0.0042 &                                        - \\
{} & $t$-SNE           &                        0.168 $\pm$ 0.031 &                 \textbf{0.404 $\pm$ 0.003} &                        0.9443 $\pm$ 0.0005 &                         39.8 $\pm$ 0.1 &                        2.9e-07 $\pm$ 4e-08 &                         0.3 $\pm$ 0.03 &                                        - \\
{} & PCA            &                 0.16276402 $\pm$ 2.2e-07 &                     0.117955 $\pm$ 1.1e-06 &                      0.7456815 $\pm$ 5e-07 &                  6.5830853 $\pm$ 8e-07 &          \textbf{1.636274e-07 $\pm$ 2e-13} &         \textbf{0.5246475 $\pm$ 7e-07} &                    0.055636764 $\pm$ 8e-09 \\

\midrule

{} & Geom AE (ours)         &   \underline{\textbf{0.0407 $\pm$ 0.0052}} &                          0.37 $\pm$ 0.03 &   \underline{\textbf{0.971 $\pm$ 0.003}} &                                7 $\pm$ 1 &                  \textbf{9.6e-08 $\pm$ 1.1e-08} &                               0.75 $\pm$ 0.03 &             \textbf{0.02562 $\pm$ 0.00013} \\
{} & Vanilla AE        &                          0.069 $\pm$ 0.031 &                          0.34 $\pm$ 0.02 &                       0.9666 $\pm$ 0.002 &                               14 $\pm$ 2 &                             1.6e-07 $\pm$ 1e-07 &                               0.66 $\pm$ 0.12 &   \underline{\textbf{0.0253 $\pm$ 0.0003}} \\
{} & Topo AE        &                  \textbf{0.049 $\pm$ 0.01} &                        0.366 $\pm$ 0.003 &                     0.9686 $\pm$ 0.00073 &                        9.569 $\pm$ 0.081 &                             1.1e-07 $\pm$ 2e-08 &                      \textbf{0.82 $\pm$ 0.02} &                        0.0261 $\pm$ 0.0002 \\
FashionMNIST & UMAP AE &                        0.0925 $\pm$ 0.0073 &                      0.4147 $\pm$ 0.0072 &               \textbf{0.971 $\pm$ 0.003} &                         10.86 $\pm$ 0.51 &                          5.36e-07 $\pm$ 3.1e-08 &                             0.595 $\pm$ 0.012 &                         0.0258 $\pm$ 0.001 \\
{} & UMAP           &                        0.0947 $\pm$ 0.0021 &               \textbf{0.422 $\pm$ 0.002} &                        0.971 $\pm$ 0.001 &   \underline{\textbf{4.416 $\pm$ 0.022}} &                            3.01e-07 $\pm$ 1e-08 &                             0.603 $\pm$ 0.002 &                                        - \\
{} & $t$-SNE           &                          0.072 $\pm$ 0.004 &   \underline{\textbf{0.441 $\pm$ 0.002}} &                     0.96872 $\pm$ 0.0006 &                             39 $\pm$ 0.2 &                           2.5e-07 $\pm$ 1.2e-08 &                               0.56 $\pm$ 0.03 &                                        - \\
{} & PCA            &                    0.052010267 $\pm$ 8e-09 &                    0.2076921 $\pm$ 4e-07 &                   0.91678396 $\pm$ 5e-08 &           \textbf{4.5253376 $\pm$ 4e-07} &   \underline{\textbf{7.084261e-08 $\pm$ 2e-14}} &   \underline{\textbf{0.88169565 $\pm$ 1e-08}} &                    0.046092747 $\pm$ 2e-09 \\

\midrule

{} & Geom AE (ours)        &   \underline{\textbf{0.047 $\pm$ 0.009}} &                         0.464 $\pm$ 0.01 &                        0.956 $\pm$ 0.002 &                          17.6 $\pm$ 1.2 &   \underline{\textbf{1.4e-07 $\pm$ 3.1e-08}} &             \textbf{0.683 $\pm$ 0.091} &                           0.73 $\pm$ 0.02 \\
{} & Vanilla AE        &                          0.09 $\pm$ 0.03 &                          0.42 $\pm$ 0.02 &                        0.943 $\pm$ 0.007 &                              36 $\pm$ 9 &                          2.8e-07 $\pm$ 1e-07 &                          0.5 $\pm$ 0.1 &                  \textbf{0.71 $\pm$ 0.02} \\
{} & Topo AE        &               \textbf{0.056 $\pm$ 0.004} &                         0.47 $\pm$ 0.007 &             \textbf{0.9561 $\pm$ 0.0013} &                          19.8 $\pm$ 0.3 &                 \textbf{1.5e-07 $\pm$ 2e-08} &   \underline{\textbf{0.72 $\pm$ 0.02}} &                         0.724 $\pm$ 0.004 \\
CElegans & UMAP AE &                        0.067 $\pm$ 0.011 &   \underline{\textbf{0.506 $\pm$ 0.004}} &   \underline{\textbf{0.963 $\pm$ 0.002}} &   \underline{\textbf{13.27 $\pm$ 0.21}} &                            2e-07 $\pm$ 6e-08 &                       0.554 $\pm$ 0.05 &   \underline{\textbf{0.6751 $\pm$ 0.004}} \\
{} & UMAP           &                        0.058 $\pm$ 0.003 &              \textbf{0.4853 $\pm$ 0.001} &                        0.946 $\pm$ 0.002 &               \textbf{13.35 $\pm$ 0.05} &                          1.6e-07 $\pm$ 8e-09 &                       0.599 $\pm$ 0.01 &                                       - \\
{} & $t$-SNE           &                        0.057 $\pm$ 0.006 &                      0.4697 $\pm$ 0.0031 &                         0.93 $\pm$ 0.004 &                          29.8 $\pm$ 0.2 &                       1.81e-07 $\pm$ 2.3e-08 &                       0.494 $\pm$ 0.02 &                                       - \\
{} & PCA            &                   0.08170186 $\pm$ 6e-08 &                 0.16197602 $\pm$ 2.1e-07 &                    0.8143107 $\pm$ 3e-07 &                  14.1533186 $\pm$ 1e-06 &                     2.501011e-07 $\pm$ 2e-13 &                  0.6426984 $\pm$ 2e-07 &                              2.06 $\pm$ 0 \\

\midrule

{} & Geom AE (ours)         &                        0.11 $\pm$ 0.013 &                      0.3945 $\pm$ 0.0072 &                 \textbf{0.943 $\pm$ 0.002} &                              17 $\pm$ 2 &                 \textbf{2.3e-07 $\pm$ 1e-07} &                               0.71 $\pm$ 0.04 &                        0.338 $\pm$ 0.004 \\
{} & Vanilla AE       &                         0.14 $\pm$ 0.01 &                        0.361 $\pm$ 0.008 &                          0.939 $\pm$ 0.003 &                              24 $\pm$ 8 &                          2.7e-07 $\pm$ 6e-08 &                               0.64 $\pm$ 0.12 &             \textbf{0.3332 $\pm$ 0.0022} \\
{} & Topo AE        &                       0.124 $\pm$ 0.003 &                        0.353 $\pm$ 0.003 &                          0.924 $\pm$ 0.003 &                        19.32 $\pm$ 0.06 &                       2.81e-07 $\pm$ 3.1e-08 &                             0.734 $\pm$ 0.021 &                      0.3431 $\pm$ 0.0004 \\
Zilionis & UMAP AE &   \underline{\textbf{0.085 $\pm$ 0.01}} &   \underline{\textbf{0.407 $\pm$ 0.002}} &   \underline{\textbf{0.9451 $\pm$ 0.0012}} &   \underline{\textbf{10.36 $\pm$ 0.07}} &                          3e-07 $\pm$ 1.2e-07 &                               0.72 $\pm$ 0.04 &   \underline{\textbf{0.332 $\pm$ 0.001}} \\
{} & UMAP           &                       0.099 $\pm$ 0.006 &                        0.387 $\pm$ 0.002 &                      0.93717 $\pm$ 0.00033 &                        12.48 $\pm$ 0.23 &                            3e-07 $\pm$ 3e-08 &                      \textbf{0.74 $\pm$ 0.03} &                                      - \\
{} & $t$-SNE           &            \textbf{0.0977 $\pm$ 0.0051} &             \textbf{0.3967 $\pm$ 0.0041} &                          0.938 $\pm$ 0.002 &                        27.09 $\pm$ 0.11 &   \underline{\textbf{2.2e-07 $\pm$ 1.2e-08}} &                              0.516 $\pm$ 0.05 &                                      - \\
{} & PCA            &                  0.11343118 $\pm$ 1e-08 &                  0.2175087 $\pm$ 3.3e-06 &                   0.86533776 $\pm$ 1.1e-07 &         \textbf{12.2618798 $\pm$ 3e-07} &                     2.944264e-07 $\pm$ 3e-13 &   \underline{\textbf{0.80789925 $\pm$ 7e-08}} &                   0.59147804 $\pm$ 2e-08 \\

\midrule

{} & Geom AE (ours)         &                   \textbf{0.0163 $\pm$ 0.0023} &   \underline{\textbf{0.2435 $\pm$ 0.0009}} &   \underline{\textbf{0.9084 $\pm$ 0.0006}} &                          6.4 $\pm$ 0.4 &               \textbf{1.1e-07 $\pm$ 2e-08} &                             0.847 $\pm$ 0.011 &   \underline{\textbf{0.3703 $\pm$ 0.0011}} \\
{} & Vanilla AE        &                            0.0653 $\pm$ 0.0032 &                          0.221 $\pm$ 0.003 &                          0.902 $\pm$ 0.002 &                             15 $\pm$ 8 &                     1.98e-07 $\pm$ 4.3e-08 &                               0.72 $\pm$ 0.09 &                 \textbf{0.371 $\pm$ 0.002} \\
{} & Topo AE        &                              0.022 $\pm$ 0.002 &                      0.23222 $\pm$ 0.00092 &               \textbf{0.9037 $\pm$ 0.0007} &                        7.37 $\pm$ 0.06 &   \underline{\textbf{7.5e-08 $\pm$ 1e-09}} &                    \textbf{0.871 $\pm$ 0.011} &                        0.3731 $\pm$ 0.0008 \\
PBMC & UMAP AE &                              0.026 $\pm$ 0.005 &               \textbf{0.2382 $\pm$ 0.0007} &                       0.90174 $\pm$ 0.0003 &                 \textbf{4.1 $\pm$ 0.3} &                        1.7e-07 $\pm$ 7e-08 &                               0.82 $\pm$ 0.01 &                      0.37955 $\pm$ 0.00081 \\
{} & UMAP           &                              0.027 $\pm$ 0.004 &                      0.21599 $\pm$ 0.00051 &                        0.8858 $\pm$ 0.0003 &   \underline{\textbf{3.84 $\pm$ 0.12}} &                     1.61e-07 $\pm$ 6.2e-08 &                               0.84 $\pm$ 0.02 &                                        - \\
{} & $t$-SNE           &                              0.038 $\pm$ 0.002 &                          0.237 $\pm$ 0.001 &                        0.8946 $\pm$ 0.0011 &                        24.4 $\pm$ 0.09 &                        1.3e-07 $\pm$ 2e-08 &                             0.674 $\pm$ 0.022 &                                        - \\
{} & PCA            &   \underline{\textbf{0.012270422 $\pm$ 3e-09}} &                     0.12920468 $\pm$ 3e-07 &                     0.82435367 $\pm$ 1e-07 &               4.91943945 $\pm$ 4.1e-07 &                  1.1969531e-07 $\pm$ 9e-14 &   \underline{\textbf{0.91108369 $\pm$ 2e-08}} &                     0.59559689 $\pm$ 3e-08 \\
\bottomrule
\end{tabularx}

\vspace{2\baselineskip}
\caption{Quantitative comparison of a convolutional vanilla autoencoder and a convolutional geometric autoencoder. Averaged over five runs, bold indicates first place. The arrows point to the desirable direction of each metric.}
\label{table:metrics-convnets}
\begin{tabularx}{\textwidth}{ll|lll|lllll}
\toprule
 \multicolumn{2}{c|}{} & \multicolumn{3}{c|}{Local} & \multicolumn{4}{c}{Global} & \\
 \midrule
Dataset & Model & $\dkl_{0.1}$ ($\downarrow$) & kNN ($\uparrow$) & Trust ($\uparrow$) & Stress ($\downarrow$) & $\dkl_{100}$ ($\downarrow$) & Spear ($\uparrow$) & MSE ($\downarrow$) \\
\midrule
{MNIST} & Conv GeomAE  &                   0.16 $\pm$ 0.01 &               0.189 $\pm$ 0.005 &              0.8396 $\pm$ 0.009 &   \textbf{4.98 $\pm$ 0.32} &               1.9e-07 $\pm$ 2e-08 &               0.52 $\pm$ 0.03 &               0.0467 $\pm$ 0.0008 \\
{} & Conv Vanilla &   \textbf{0.1304 $\pm$ 0.0043} &   \textbf{0.195 $\pm$ 0.005} &   \textbf{0.847 $\pm$ 0.005} &                     6 $\pm$ 2 &   \textbf{1.7e-07 $\pm$ 4e-08} &   \textbf{0.57 $\pm$ 0.03} &   \textbf{0.0456 $\pm$ 0.0005} \\
\midrule
{FashionMNIST} & Conv GeomAE (ours)  &               0.051 $\pm$ 0.005 &   \textbf{0.283 $\pm$ 0.004} &   \textbf{0.951 $\pm$ 0.0008} &   \textbf{5.27 $\pm$ 0.92} &   \textbf{9.7e-08 $\pm$ 1.3e-08} &               0.82 $\pm$ 0.03 &              0.0322 $\pm$ 0.0004 \\
{} & Conv Vanilla &   \textbf{0.046 $\pm$ 0.006} &             0.2809 $\pm$ 0.0021 &            0.95089 $\pm$ 0.00053 &                14.3 $\pm$ 1.1 &                 1.1e-07 $\pm$ 2e-08 &   \textbf{0.85 $\pm$ 0.02} &   \textbf{0.0312 $\pm$ 0.0002} \\
\bottomrule
\end{tabularx}

\end{sc}
}
\end{small}
\end{center}
\vskip -0.1in
\end{sidewaystable}

\end{document}